\pdfoutput=1
\documentclass[]{sfuthesis}

\usepackage[english]{babel}
\usepackage[T1]{fontenc}
\usepackage[utf8]{inputenc}
\usepackage{newtxtext}

\usepackage[hidelinks, pdfusetitle]{hyperref}

\usepackage{amssymb}
\usepackage[titletoc]{appendix}
\usepackage[style=ieee]{biblatex}
\usepackage[font=small, skip=12pt, labelfont=bf]{caption}
\usepackage{csquotes}
\usepackage{fancyvrb}
\usepackage{float}
\usepackage{fp}
\usepackage{geometry}
\usepackage{graphicx}
\usepackage{grffile}
\usepackage{import}
\usepackage{makecmds}
\usepackage{mathtools}
\usepackage[frozencache=true,cachedir=minted-cache]{minted}
\usepackage[super]{nth}
\usepackage{placeins}
\usepackage{setspace}
\usepackage{standalone}
\usepackage{subcaption}
\usepackage{tabularx}
\usepackage{titlesec}
\usepackage{trimspaces}
\usepackage{xcolor}

\usepackage{tikz}

\usepackage{alphalph}

\usepackage[capitalize, nameinlink, noabbrev]{cleveref}

\provideenvironment{abstract}{}{}
\usepackage{abstract}

\usepackage{review}

\usetikzlibrary{
  arrows.meta,
  calc,
  external,
  hobby,
  shapes,
}

\newgeometry{
  top=1in,
  left=1.25in,
  bottom=0.7in,
  right=1.25in,
  includefoot
}

\MakeOuterQuote{"}

\DeclareFieldFormat{title}{\mkbibquote{#1}}

\makeatletter
\newcommand*{\trim}[1]{%
  \trim@spaces@noexp{#1}%
}
\makeatother

\newenvironment{conditions}{%
  \par\vspace{\abovedisplayskip}\noindent
  \tabularx{\columnwidth}{%
    >{$}l<{$} @{${}={}$} >{\raggedright\arraybackslash}X%
  }%
}{%
  \endtabularx\par\vspace{\belowdisplayskip}%
}

\newcommand{\hwc}[3]{%
  $#1 \times #2 \times #3$%
}

\newcommand{\csubref}[1]{%
  (\subref{#1})%
}

\crefname{equation}{}{}

\newif\ifhighlightcode
\highlightcodetrue

\makeatletter
\ifhighlightcode%

\else%
  \newenvironment{codelisting}[2][]{%
    \VerbatimEnvironment
    \begin{Verbatim}%
  }{%
    \end{Verbatim}%
    \ignorespacesafterend%
  }%
\fi
\makeatother

\title{Mobile-Cloud~Inference for Collaborative~Intelligence}
\author{Mateen Ulhaq}
\date{June 9, 2020}

\thesistype{Thesis}
\previousdegrees{}
\degree{Bachelor of Applied Science (Honours)}
\discipline{Engineering Physics}
\department{School of Engineering Science}
\faculty{Faculty of Applied Sciences}
\copyrightyear{2020}
\semester{Spring 2020}


\committee{%
  \chair{Dr. Ivan V. Baji\'c, P.Eng.}{Academic Supervisor \\ Professor}
  \member{Dr. Mirza Faisal Beg, P.Eng.}{Committee \\ Professor}
  \member{Dr. Jie Liang, P.Eng.}{Committee \\ Professor}
}

\begin{document}

\frontmatter

\maketitle
\makecommittee
\clearpage
\phantomsection
\addcontentsline{toc}{chapter}{Abstract}
\newcounter{abstractpage}
\setcounter{abstractpage}{\value{page}}

\begin{abstract}
  \thispagestyle{plain}
  \setcounter{page}{\value{abstractpage}}

  As AI applications for mobile devices become more prevalent, there is an
  increasing need for faster execution and lower energy consumption for deep
  learning model inference. Historically, the models run on mobile devices have
  been smaller and simpler in comparison to large state-of-the-art research
  models, which can only run on the cloud. However, cloud-only inference has
  drawbacks such as increased network bandwidth consumption and higher latency.
  In addition, cloud-only inference requires the input data (images, audio) to
  be fully transferred to the cloud, creating concerns about potential privacy
  breaches.

  There is an alternative approach: shared mobile-cloud inference. Partial
  inference is performed on the mobile in order to reduce the dimensionality of
  the input data and arrive at a compact feature tensor, which is a latent
  space representation of the input signal. The feature tensor is then
  transmitted to the server for further inference. This strategy can reduce
  inference latency, energy consumption, and network bandwidth usage, as well
  as provide privacy protection, because the original signal never leaves the
  mobile. Further performance gain can be achieved by compressing the feature
  tensor before its transmission.

  \setcounter{abstractpage}{\value{page}}
\end{abstract}

\setcounter{page}{\value{abstractpage}}
\stepcounter{page}

\chapter*{Acknowledgments}
\addcontentsline{toc}{chapter}{Acknowledgements}

Firstly, I would like to thank my academic supervisor Dr. Ivan V. Baji\'c for
his guidance, patience, and support. His feedback in reviewing and editing my
thesis was invaluable. His experience and thoughtful ideas during meetings
helped me navigate this new area of research. Furthermore, his encouragement
motivated me to develop a prototype for demonstration at the NeurIPS 2019
conference.

I would also like to thank my committee members. Dr. Mirza Faisal Beg, who was
my supervisor for a NSERC Undergraduate Student Research Awards (USRA) research
project, and provided me with an opportunity to work on a challenging problem
in computer vision, related to medical mobile applications. Dr. Jie Liang, who
inspired and intrigued me with his lectures on Multimedia Communications
Engineering involving the subjects of compression and deep learning.

The assistance my lab members Alon Harell and Hyomin Choi provided me during my
demonstration at the NeurIPS 2019 conference was helpful in ensuring everything
ran smoothly. Furthermore, their comments, along with those of Dr. Robert A.
Cohen and Tim Woinoski, helped shaped the presentation of the demo and my
thesis.

I would also like to show my appreciation for the community at SFU, including
my friends and peers, as well as the professors of the schools of engineering
science and computer science, and the departments of physics and mathematics.

And of course, for the neverending support from my family, including my mother
and sister, I am grateful.

\clearpage
\phantomsection
\pdfbookmark{\contentsname}{toc}
\tableofcontents
\listoffigures

\clearpage
\pagenumbering{arabic}

\mainmatter

\chapter{Introduction}
\label{chap:introduction}

Deep learning models have expanded the capabilities of mobile devices. Common
applications include speech recognition, face recognition, pose estimation,
object detection, night vision, artistic style transfer, and other image
analysis and modification capabilities. However, more powerful and useful
models also come with a significant computational load that takes time to run,
and may be a drain on resources such as energy or the amount of available
processing power for other computational tasks.

\begin{figure}
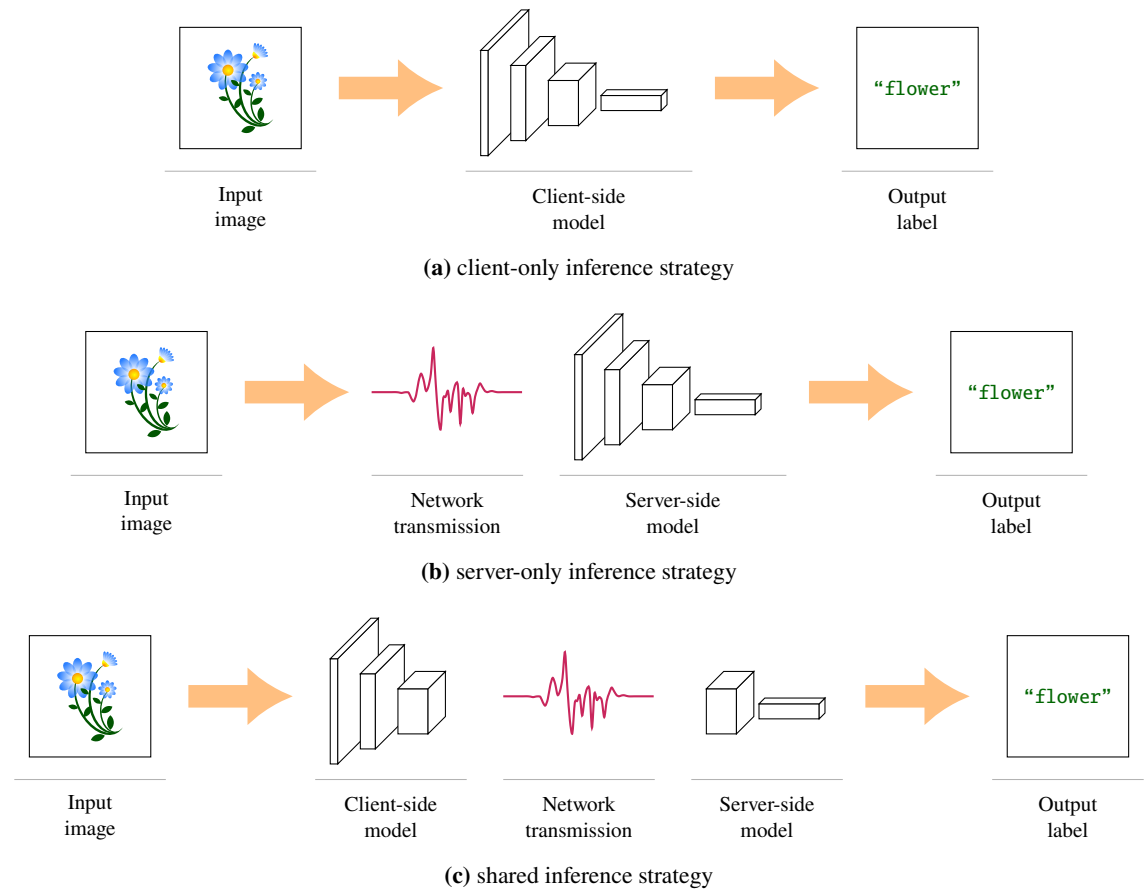

  \centering

  \begin{subfigure}[t]{\linewidth}
    \centering
    \import{tikz/strategies/}{client.tex}
    \caption{%
      client-only inference strategy%
    }
    \label{fig:strategies/client}
  \end{subfigure}

  \begin{subfigure}[t]{\linewidth}
    \centering
    \import{tikz/strategies/}{server.tex}
    \caption{%
      server-only inference strategy%
    }
    \label{fig:strategies/server}
  \end{subfigure}

  \begin{subfigure}[t]{\linewidth}
    \centering
    \import{tikz/strategies/}{shared.tex}
    \caption{%
      shared inference strategy%
    }
    \label{fig:strategies/shared}
  \end{subfigure}

  \caption[Comparison of inference strategies]{%
    Comparison of inference strategies. %
    \csubref{fig:strategies/client} Client-only inference strategy. Advantages:
    no data is transmitted over the network. Disadvantages: client-side models
    may be limited in size and complexity, consume energy, or take up
    computational resources away from other tasks. %
    \csubref{fig:strategies/server} Server-only inference strategy. Advantages:
    ability to run heavier and more powerful models or multi-model.
    Disadvantages: may consume large amounts of bandwidth, susceptible to
    network latencies and instability, and privacy concerns from sending input
    data. %
    \csubref{fig:strategies/shared} Shared inference strategy.
    The input image is first processed through the client-side model. The
    output of the client-side model is then serialized and transmitted over the
    network. The server receives the data, deserializes it, and then feeds it
    into the server-side model. Finally, the server-side model outputs the
    desired result. Advantages: allows for more computationally expensive
    inference models, potentially reduced bandwidth usage in comparison to
    server-only inference, and better privacy protection. Disadvantages: not
    suitable for all model architectures.%
  }
  \label{fig:strategies}
\end{figure}

A client-only inference strategy, shown in \cref{fig:strategies/client},
performs all processing on the client itself. In order to reduce the effect of
the aforementioned negative factors on the mobile device, one may attempt to
offload the computation by performing the complete inference on the server.
This involves sending the input signal (e.g. images or audio) to the server
over a network. This server-only inference strategy is visualized in
\cref{fig:strategies/server}. However, the amount of data sent to the server
can consume large amounts of bandwidth, time, and energy, depending on the
quality of the network connection. Also, sending images or audio directly to
the cloud may raise privacy concerns.

This thesis considers a different approach: shared inference, also known as
\emph{collaborative intelligence}~\cite{neurosurgeon,JointDNN,choi_icip_2018}.
In an effort to reduce the inference latency and the amount of data sent over
the network, the first part of the inference is performed on the client. The
resulting intermediate tensor%
\footnote{%
  In machine learning and related fields, the word "tensor" is typically used
  to describe a multidimensional array. A tensor $T$ can be \emph{indexed} in
  the same way as an array. The number of indices required to reduce the tensor
  to a single number corresponds to its dimension. For instance, a vector is a
  1-d tensor indexed by $T_i$; a matrix is a 2-d tensor indexed by $T_{ij}$; a
  "cube" of numbers is a 3-d tensor indexed by $T_{ijk}$; and so on.%
}
is then compressed, serialized, and sent over the network. The server then
reconstructs an approximation of the original tensor and completes the second
part of the inference. The result can be transmitted back to the client and/or
used for other analytics in the cloud. This is visualized in
\cref{fig:strategies/shared}. In addition to possible energy savings and
improved latency~\cite{neurosurgeon,JointDNN}, collaborative intelligence
offers better possibilities for privacy protection because the original input
signal never leaves the client device. What is transmitted to the cloud is the
latent space representation of the input signal in the form of one or more deep
feature tensors.

\section{On compressibility of deep features}
\label{sec:introduction/compressibility}

To motivate why collaborative intelligence may be feasible in terms of the
amount of data that needs to be transferred to the server, it first helps to
think about things from the point of view of information theory. By the
\emph{data processing inequality}~\cite{cover2006elements}, the mutual
information between a signal and its processed version decreases as more and
more processing is performed. Specifically, if $X$, $Y$, and $Z$ are random
variables (or vectors) and  if $X \rightarrow Y \rightarrow Z$ is a Markov
chain, then $I(X; Z) \leq I(X; Y)$. The inequality is non-strict, which means
that mutual information need not decrease --- it could stay the same. Data
processing in a deep learning model with $L$ layers may be formulated~%
\cite{zhang2018informationtheoretic,weng2017infotheory,tishby2015deep} as a
Markov chain $X_1 \to X_2 \to \ldots \to X_L$, where $X_1$ is the input, and
$X_l$ is the output of the $l$-th layer. Then, the data processing inequality
implies that
\begin{equation}
  I(X_1; X_2) \geq I(X_1; X_3) \geq \ldots \geq I(X_1; X_L).
  \label{eq:data_proc_ineq}
\end{equation}
Thus, mutual information between the input $X_1$ and the outputs of successive
layers is non-increasing.

The above reasoning only says that information \emph{about the input} $X_1$
cannot increase as we move through the layers of a deep model; it does not say
that the amount of information (i.e. the entropy) of $X_l$ decreases as $l$
increases. If we want to compress $X_l$, we need to examine the entropy of
$X_l$, denoted $H(X_l)$. For this purpose, we recall another result from
information theory~\cite{cover2006elements}:
\begin{equation}
    I(X_1; X_l) = H(X_l) - H(X_l~|~X_1).
    \label{eq:MI_entropy}
\end{equation}
The last term in \cref{eq:MI_entropy} is the conditional entropy of $l$-th
layer's output $X_l$ given the input $X_1$; it is the uncertainty in the value
of $X_l$, given $X_1$. In non-generative feedforward neural networks, each
layer's output is a deterministic function of the input. That is to say, when a
given input $X_1$ is presented to the network, each layer's output is uniquely
determined through feedforward propagation. There is no uncertainty in the
value of $X_l$, given $X_1$, and therefore $H(X_l~|~X_1) = 0$. Combined with
the data processing inequality, this leads to the conclusion that in
non-generative feedforward networks,
\begin{equation}
    H(X_1) \geq H(X_2) \geq \ldots \geq H(X_L).
    \label{eq:entropy_decreasing}
\end{equation}

Since the entropy is the limit of (lossless)
compressibility~\cite{cover2006elements}, \cref{eq:entropy_decreasing} predicts
that the features taken from inside a non-generative feedforward network are at
least as compressible as the input; and they become no less compressible as we
move deeper into the network. This provides a theoretical basis for
collaborative intelligence achieving the goal of sending less data to the
server, compared to sending the input signal. Of course, this way of thinking
merely provides theoretical bounds on information content and compressibility;
it does not tell us how to construct a practical codec to achieve these limits.

\begin{figure}
  \centering
  \tikzstyle{block} = [
  rectangle,
  draw,
]

\tikzstyle{cutvx} = [
  rectangle,
  draw,
  red,
]

\tikzstyle{terminal} = [
  rectangle,
  draw,
]

\tikzstyle{line} = [draw, thick, -latex]
\tikzstyle{bendleft} = [bend left=15]
\tikzstyle{bendright} = [bend right=15]
\tikzstyle{cutvertexcut} = [semithick, dashed]
\tikzstyle{stcut} = [thick, dotted]
\tikzstyle{hobbycut} = [stcut, use Hobby shortcut, out angle=0, in angle=180]
\tikzstyle{ta} = [tension in=2]
\tikzstyle{tb} = [tension out=2]

\begin{tikzpicture}[node distance=1.0cm, auto]
  \node[terminal] (input) {Input};
  \node[cutvx, below of=input] (A) {BN};
  \node[cutvx, below of=A] (B) {Conv};
  \node[cutvx, below of=B] (C) {ReLU};
  \node[block, below of=C, shift={(180:1)}] (d1_L1) {Conv};
  \node[block, below of=C, shift={(0:1)}, shift={(-90:1)}] (d1_R1) {Conv};
  \node[block, below of=d1_L1] (d1_L2) {ReLU};
  \node[block, below of=d1_L2] (d1_L3) {Conv};
  \node[cutvx, below of=d1_L3, shift={(0:1)}] (add1) {Add};
  \node[block, below of=add1, shift={(180:1)}] (d2_L1) {Conv};
  \node[block, below of=add1, shift={(0:1)}, shift={(-90:1)}] (d2_R1) {Conv};
  \node[block, below of=d2_L1] (d2_L2) {ReLU};
  \node[block, below of=d2_L2] (d2_L3) {Conv};
  \node[cutvx, below of=d2_L3, shift={(0:1)}] (add2) {Add};
  \node[cutvx, below of=add2] (fc) {FC};
  \node[terminal, below of=fc] (output) {Output};

  \path[line] (input) -- (A);
  \path[line] (A) -- (B);
  \path[line] (B) -- (C);
  \path[line] (C) edge [bendright] (d1_L1);
  \path[line] (C) edge [bendleft] (d1_R1);
  \path[line] (d1_L1) -- (d1_L2);
  \path[line] (d1_L2) -- (d1_L3);
  \path[line] (d1_L3) edge [bendright] (add1);
  \path[line] (d1_R1) edge [bendleft] (add1);
  \path[line] (add1) edge [bendright] (d2_L1);
  \path[line] (add1) edge [bendleft] (d2_R1);
  \path[line] (d2_L1) -- (d2_L2);
  \path[line] (d2_L2) -- (d2_L3);
  \path[line] (d2_L3) edge [bendright] (add2);
  \path[line] (d2_R1) edge [bendleft] (add2);
  \path[line] (add2) -- (fc);
  \path[line] (fc) -- (output);

  \draw[cutvertexcut] (-2, -1.5)  -- (2, -1.5);
  \draw[cutvertexcut] (-2, -2.5)  -- (2, -2.5);
  \draw[cutvertexcut] (-2, -3.5)  -- (2, -3.5);
  \draw[cutvertexcut] (-2, -7.5)  -- (2, -7.5);
  \draw[cutvertexcut] (-2, -11.5) -- (2, -11.5);
  \draw[cutvertexcut] (-2, -12.5) -- (2, -12.5);

  \draw[stcut]        (-2, -0.5)  -- (2, -0.5);

  \draw[hobbycut, yshift=-4cm]
    (-2, -0.5) .. ([ta]-0.5, -0.4) .. (0, -0.25) .. ([tb]0.5, -0.1) .. (2, 0);
  \draw[hobbycut, yshift=-4cm]
    (-2, -1.5) .. ([ta]-0.5, -1.4) .. (0, -0.75) .. ([tb]0.5, -0.2) .. (2, 0);
  \draw[hobbycut, yshift=-4cm]
    (-2, -2.5) .. ([ta]-0.5, -2.4) .. (0, -1.25) .. ([tb]0.5, -0.3) .. (2, 0);
  \draw[hobbycut, yshift=-4cm]
    (-2, -0.5) .. ([ta]-0.5, -0.6) .. (0, -1.25) .. ([tb]0.5, -1.8) .. (2, -2);
  \draw[hobbycut, yshift=-4cm]
    (-2, -1.5) .. ([ta]-0.5, -1.6) .. (0, -1.75) .. ([tb]0.5, -1.9) .. (2, -2);
  \draw[hobbycut, yshift=-4cm]
    (-2, -2.5) .. ([ta]-0.5, -2.4) .. (0, -2.25) .. ([tb]0.5, -2.1) .. (2, -2);

  \draw[hobbycut, yshift=-8cm]
    (-2, -0.5) .. ([ta]-0.5, -0.4) .. (0, -0.25) .. ([tb]0.5, -0.1) .. (2, 0);
  \draw[hobbycut, yshift=-8cm]
    (-2, -1.5) .. ([ta]-0.5, -1.4) .. (0, -0.75) .. ([tb]0.5, -0.2) .. (2, 0);
  \draw[hobbycut, yshift=-8cm]
    (-2, -2.5) .. ([ta]-0.5, -2.4) .. (0, -1.25) .. ([tb]0.5, -0.3) .. (2, 0);
  \draw[hobbycut, yshift=-8cm]
    (-2, -0.5) .. ([ta]-0.5, -0.6) .. (0, -1.25) .. ([tb]0.5, -1.8) .. (2, -2);
  \draw[hobbycut, yshift=-8cm]
    (-2, -1.5) .. ([ta]-0.5, -1.6) .. (0, -1.75) .. ([tb]0.5, -1.9) .. (2, -2);
  \draw[hobbycut, yshift=-8cm]
    (-2, -2.5) .. ([ta]-0.5, -2.4) .. (0, -2.25) .. ([tb]0.5, -2.1) .. (2, -2);
\end{tikzpicture}
  \caption[Deep learning model as a graph]{%
    Deep model with both single-stream and multi-stream sections. Dashed and
    dotted lines show various possibilities for cutting the model. Cuts along
    dashed lines result in one feature tensor, those along dotted lines result
    in two feature tensors.%
  }
  \label{fig:graph_cut}
\end{figure}
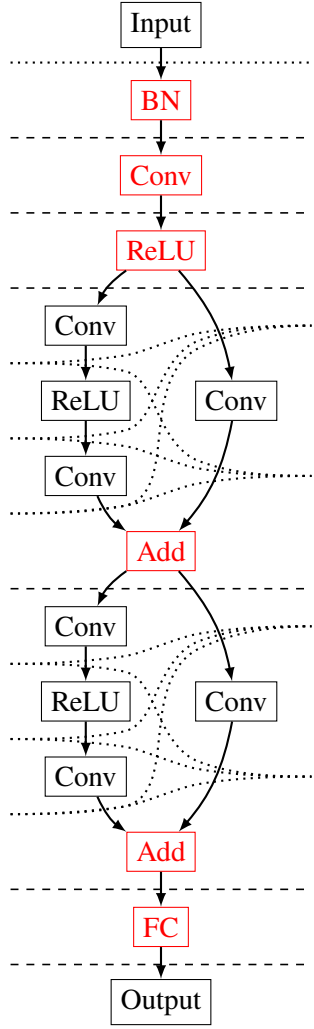

More generally, consider a set of outputs of $n$ layers in a deep
non-generative model: $\{X_{j_1}, X_{j_2}, \ldots, X_{j_n}\}$. These could be
outputs of various layers in a single-stream model like VGG16~\cite{VGG}, or a
multi-stream model like U-Net~\cite{U-Net}. Some models, such as
ResNet~\cite{he2016deep}, have both single-stream and multi-stream sections. A
ResNet-like structure is shown in \cref{fig:graph_cut} for illustration.

Let $Y = (X_{j_1}, X_{j_2}, \ldots, X_{j_n})$ be their joint output. Note that
$Y$ is a deterministic function of the input $X_1$ --- when $X_1$ is given, $Y$
is uniquely determined, hence $H(Y~|~X_1) = 0$. By considering the Markov chain
$X_1 \to X_1 \to Y$ and using the data processing inequality, we can write
\begin{equation}
  \begin{split}
    H(X_1) = I(X_1; X_1)
    &\geq I(X_1; Y) \\
    &= H(Y) - H(Y~|~X_1) \\
    &= H(Y) \\
    &= H(X_{j_1}, X_{j_2}, \ldots, X_{j_n}).
  \end{split}
  \label{eq:entropy_multi-stream}
\end{equation}
Hence, the joint entropy of the outputs of $n$ layers in a non-generative model
is no larger than the entropy of the input. According to this result, in the
limit, we should be able to compress (at least losslessly) outputs from any
number of internal layers of a non-generative model, so long as we do it
jointly. Again, as with the result in \cref{eq:entropy_decreasing}, this is a
statement about limits of compression performance; it does not tell us how to
actually construct such a joint codec.

The above discussion applies to lossless compression of input and deep feature
tensors. Analogous results for lossy compression are yet to be obtained,
however, we believe that these also hold because in non-generative models, all
layer outputs are uniquely determined once the input is presented to the model.
The work presented in this thesis focuses on single-stream cuts, where one
input produces a single tensor at the cut point. We study the performance of
practical lossy compression for such scenarios.

\section{Choosing where to cut a model}
\label{sec:introduction/choosing_cut}

The best point to split a model at depends on a variety of factors. Selection
of the best cut point has been studied by several research groups.
In~\cite{neurosurgeon}, the authors empirically select the cut points of
various deep models based on measuring energy consumption at the mobile device
and/or end-to-end latency. Their results indicate that the best cut point
depends on the type of connectivity between the mobile and the cloud. If the
mobile is connected via a Wi-Fi connection (high-bitrate, close to access
point), the optimal cut point tends to be closer to the input, sometimes the
input itself (i.e. sending the input signal directly). For cellular connections
such as 3G or LTE (lower bitrate, far from access point), the optimal cut point
is deeper in the model. The primary reason is that deeper cut points result in
smaller data volumes, which results in less energy being used for the radio.
Another study~\cite{JointDNN} presents a more comprehensive model of resource
utilization in collaborative intelligence, which allows cut points to be
selected by considering not only energy utilization at the mobile and
end-to-end latency, but also cloud congestion and Quality of Service (QoS).

\begin{figure}
  \centering
  \begin{subfigure}[t]{0.85\linewidth}
    \centering
    \includegraphics[width=\linewidth]{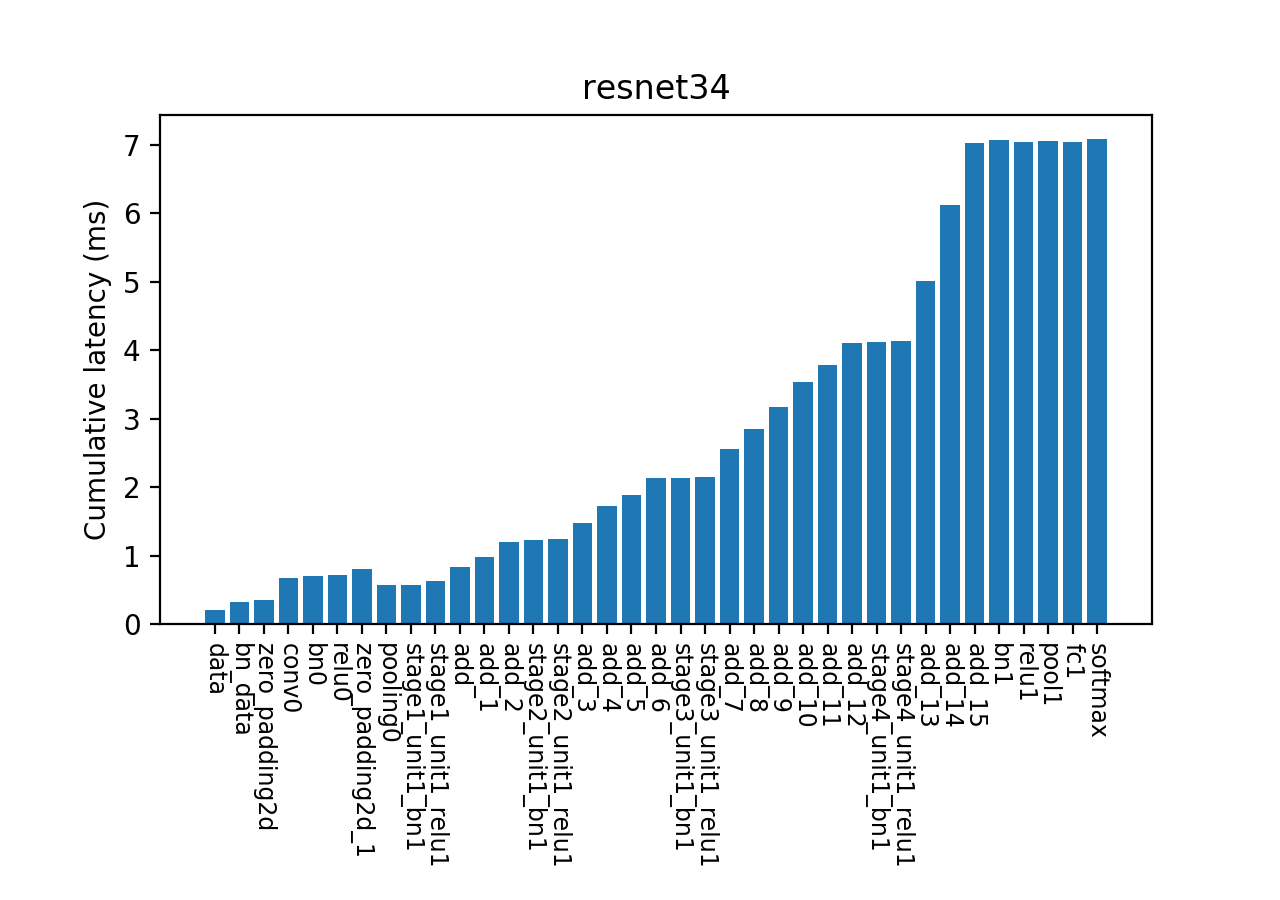}
    \caption{%
      cumulative inference latencies%
    }
    \label{fig:surgeon/latency}
  \end{subfigure}

  \begin{subfigure}[t]{0.85\linewidth}
    \centering
    \includegraphics[width=\linewidth]{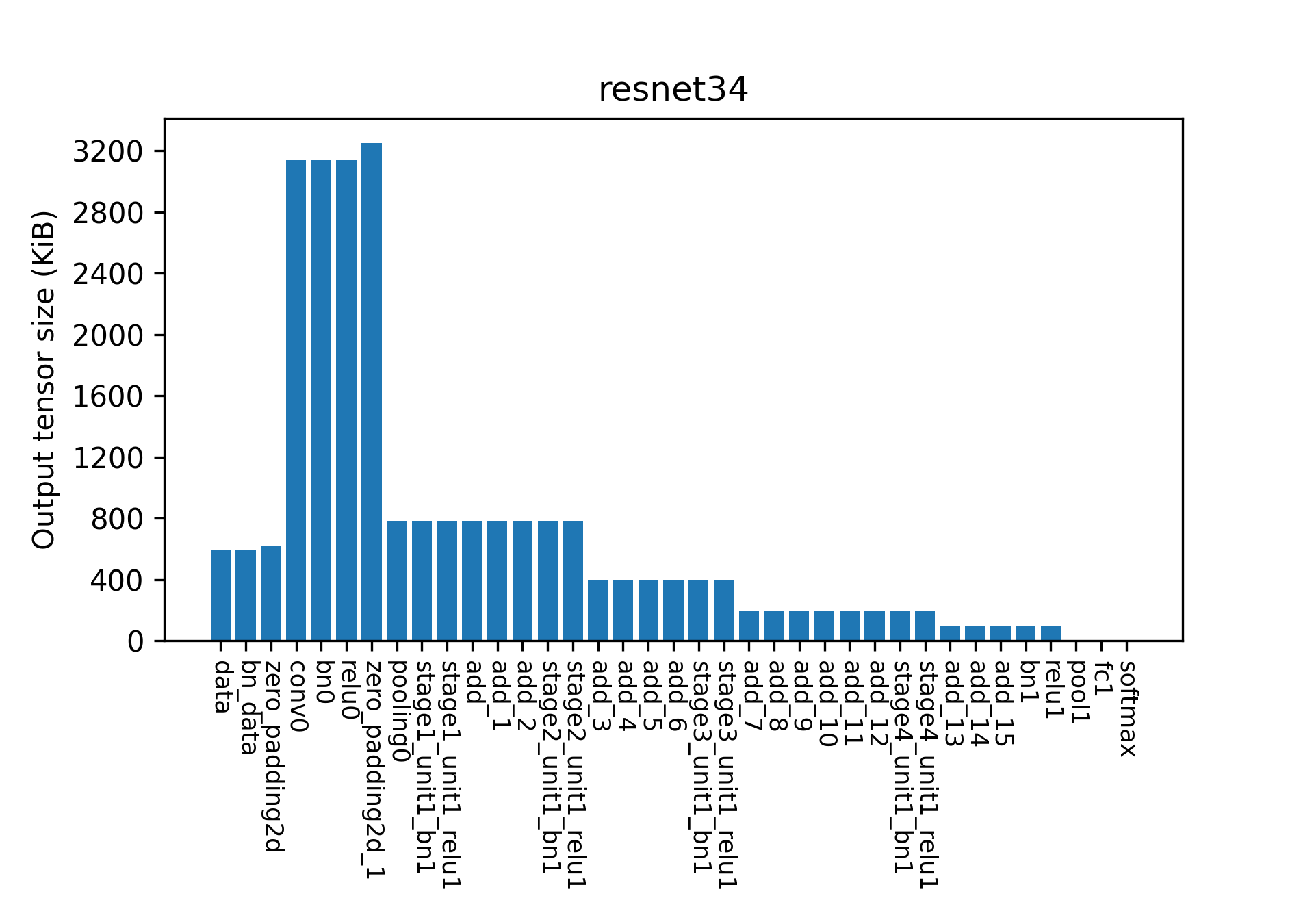}
    \caption{%
      output tensor sizes%
    }
    \label{fig:surgeon/size}
  \end{subfigure}

  \caption[Layer inference latencies and output tensor sizes at cut points]{%
    Layer inference latencies and output tensor sizes of ResNet-34 at cut
    points. The cumulative inference latencies were determined by running
    inference on the model up until the given cut point on an NVIDIA GeForce
    GTX Titan X. Note that the cumulative inference latencies are not
    necessarily monotonically increasing as we move deeper into the model; this
    is likely due to the reduced cost of copying tensors to output nodes (e.g.
    to a separate VRAM buffer) as their sizes decrease. The transfer latency
    between GPU and CPU (or rather, VRAM and RAM) is not included here.%
  }
  \label{fig:surgeon}
\end{figure}

Consider \cref{fig:surgeon/latency}, which plots the cumulative computation
time%
\footnote{%
  The cumulative computation time at a particular cut point refers to the
  amount of time it takes for inference to complete from the input to the cut
  point.%
}
at various cut points, ordered in distance from the source. As expected, the
cumulative computation time at cut points is monotonically increasing. By
splitting the model at as early a cut point as possible, we can keep the amount
of computation performed by the client-side model as small as possible and
offload the rest of the computation to the server. However, if the split is
performed too early in the model, the amount of information contained in the
signal may remain too large to compress effectively. The output tensor sizes
plotted in \cref{fig:surgeon/size} provide an upper bound on the compressed
tensor sizes. Thus, models best suited towards a shared inference strategy are
those that have a small computational footprint and reduce the input dimensions
significantly in the early layers. Of course, due to the possibility of
compression, these factors merely provide a rough guide.

\begin{figure}
  \centering
  \includegraphics[width=\linewidth]%
    {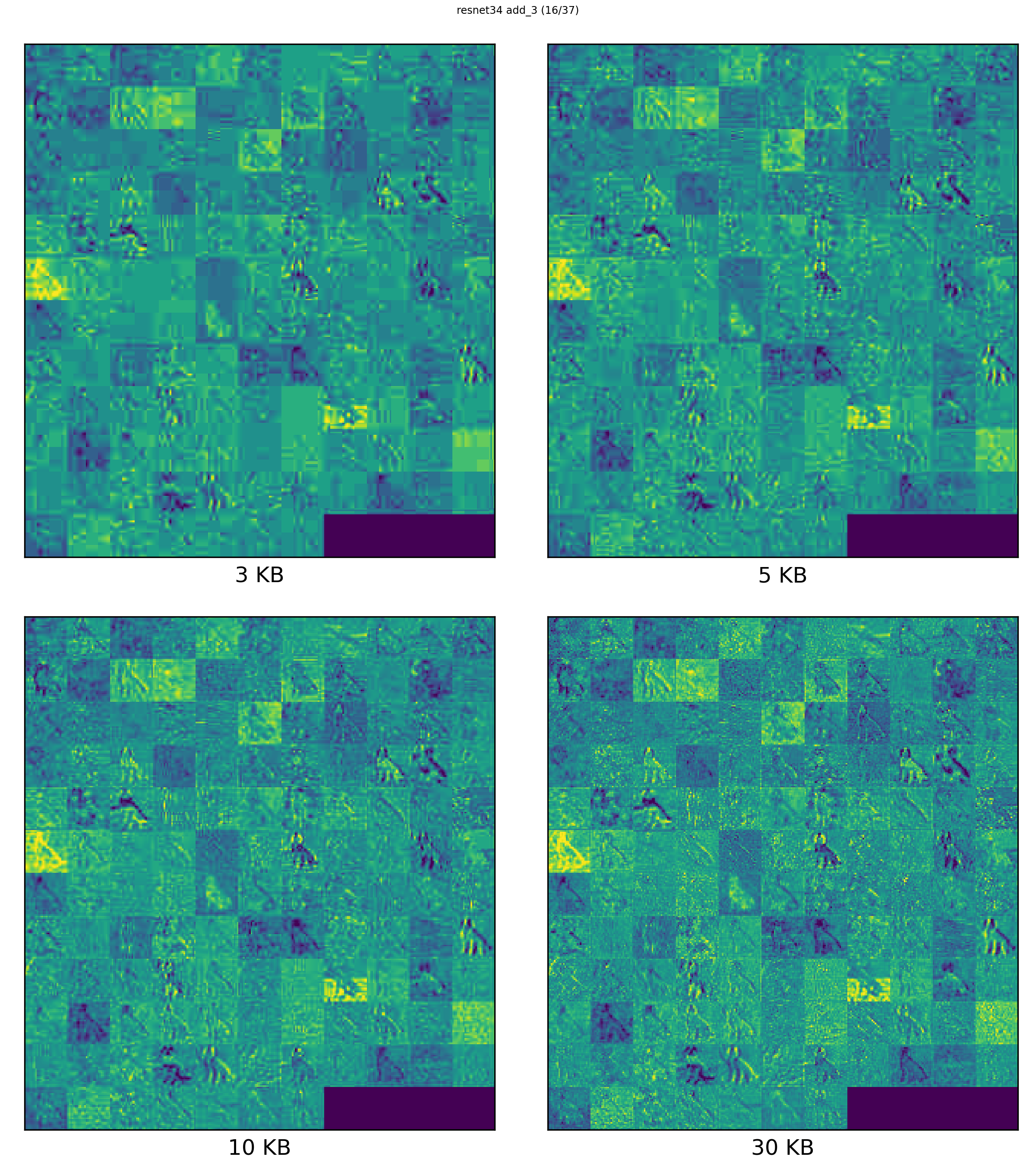}
  \caption[Feature map of intermediate layer at different compression levels]{%
    Feature map of intermediate \texttt{add\_3} layer of ResNet-34 compressed
    to different sizes. The tensor of shape $(h, w, c)$ is uniformly quantized
    to 8-bits, then tiled into a grid of $(h, w)$ shaped channels. Different
    levels of JPEG compression are applied to the resulting "image". Shown here
    are JPEGs of size 3, 5, 10, and 30 KiB. (The last of these is visually
    identical to the original tensor "image".) The preceding layers within the
    network consist of conv2d layers, pooling layers, BatchNorm layers,
    addition layers, and ReLU activation layers. Because of this, some
    characteristics of the original image are still visually apparent within
    the individual channels (e.g. the outline of a dog). There are also
    inter-channel similarities. For example, one can classify the channels as
    "light", "dark", "noisy", "smooth", or some combination thereof. These
    inter-channel similarities can be exploited for better compression.%
  }
  \label{fig:tensor_compression_levels}
\end{figure}

The output tensors of intermediate layers of convolutional neural networks
(CNNs) can often have structural redundancies between channels, which can be
exploited for heavy compression. Consider \cref{fig:tensor_compression_levels},
which shows a tiled layout of the channels within the intermediate layer of a
CNN. Within a given channel, nearby pixels often have similar intensity values.
Furthermore, there also exist structural similarities across channels. These
properties of local similarity and inter-channel similarity can be exploited to
compress the data.

One also needs to consider how the compression scheme affects the model's
overall inference accuracy. While lossless compression is possible, it is often
not very practical. Thus, we will be looking at lossy compression schemes that
result in imperfect reconstructions of the intermediate signal, which can have
an effect on the model's accuracy.

In many applications, one is often performing inference on a series of inputs.
These inputs often share structural similarities with each other. For instance,
a series of frames taken from a video often have temporal redundancies that are
exploited by video codecs. Thus, it is useful to also produce compression
schemes for tensor "streams". One desirable property of the input tensor for
such a compression scheme is the stability of the tensor values with respect to
minor changes in the input. For example, it is desirable that a small
translation of the input image should result in a small recognizable change in
the output tensor.

To summarize, when choosing which cut point to split the model at, one needs to
make the following considerations:

\begin{itemize}
  \item Is there a good split point that can be chosen as close to the input as
    possible, to minimize the cumulative computation from preceding layers?
  \item How compressible is the split layer's output tensor data?
  \item How is inference accuracy affected by the lossy compression scheme
    applied to the split layer's output tensor data?
  \item How stable is the split layer's tensor output data with respect to
    minor changes in input data?
\end{itemize}

\section{Thesis preview}

In this thesis, we will look at developing tensor compression techniques,
improving them, and also consider the challenges in developing a working
implementation of collaborative intelligence. In \cref{chap:compression}, we
will begin developing compression techniques for tensor data while attempting
to maintain a good overall inference accuracy. \cref{chap:tensor_streams} will
demonstrate useful properties that can be used to develop better compression of
tensor streams as well as the handling of missing or corrupted tensor data.
Finally, in \cref{chap:implementation}, we will discuss a proof-of-concept
Android and server implementation for shared inference that was demonstrated at
the NeurIPS 2019 conference~\cite{ulhaq2020shared}; we will also discuss other
challenges that come up in creating real working implementations of
collaborative intelligence and what can be done to resolve them.

\chapter{Single tensor compression}
\label{chap:compression}

One of the primary determinants for feasibility of the shared inference
strategy is the amount of data transmitted over the network. The less data that
is transmitted, the better the shared inference strategy performs in comparison
to the server-only inference strategy. Common sources of input data such as
images, video, and audio often have mature codecs that offer reasonable amounts
of compression at low computational cost (albeit, with some privacy concerns).
In order to be a competitive alternative to server-only inference, a shared
inference strategy needs to be on par or outperform compression of input data.
Thus, a major topic for mobile-cloud shared inference is the compression of
intermediate tensor data. It is important to keep in mind that the primary
objective is achieving good \emph{inference accuracy} rather than reducing
reconstruction error, though these goals often do coincide.

In this chapter, we will be investigating layer properties, quantization
methods, and techniques for single tensor compression. First, we will see that
models with BatchNorm layers often provide good candidate layers to split upon
due to the light-tailed shape of their output distributions. Upon the output of
these layers, we will apply quantization with varying levels and bins, and see
how this affects the final inference accuracy. Then, we will look at specific
methods for compressing single tensors from CNNs, for which we will utilize
existing codecs such as JPEG and JPEG 2000.

\section{Quantization}

A first step towards good compression is quantization of the floating-point
values of the tensor data. Though lossless compression of floating-point values
themselves is possible, much of the precision they provide often does not
affect accuracy significantly. Indeed, as we will soon see, one may sometimes
get away with as few as 2 bits of precision per tensor element.

\subsection{BatchNorm}

Many models contain BatchNorm~\cite{ioffe2015batch} layers to assist in
training. It is a popularly held belief that training is sped up due to the
reduction of Internal Covariate Shift (ICS) that occurs at a deep layer as a
result of changing the parameters of earlier layers; however, this has been
shown to be false~\cite{santurkar2018does}. Nonetheless, BatchNorm layers do
appear to help limit the possible range of neuron output values by
"normalizing" the values by their standard deviation within mini-batches. By
keeping a running average of the mean and variance, the estimated mean and
variance also apply to a larger variety of inputs.

The BatchNorm equations are reproduced as follows~\cite{ioffe2015batch}:
\begin{align}
  \mu_B &= \frac{1}{m} \sum_{i=1}^m x_i \\
  \sigma_B^2 &= \frac{1}{m} \sum_{i=1}^m (x_i - \mu_B)^2 \\
  \hat{x}_i &= \frac{x_i - \mu_B}{\sqrt{\sigma_B^2 + \epsilon}} \\
  y_i &= \textrm{BN}_{\gamma, \beta}(x_i)
    = \gamma \hat{x}_i + \beta
\end{align}
where $B$ refers to a mini-batch of size $m$, and $\gamma$ and $\beta$ are
trainable parameters.

\begin{figure}
  \centering
  \begin{subfigure}{.5\textwidth}
    \centering
    \includegraphics[width=\linewidth]{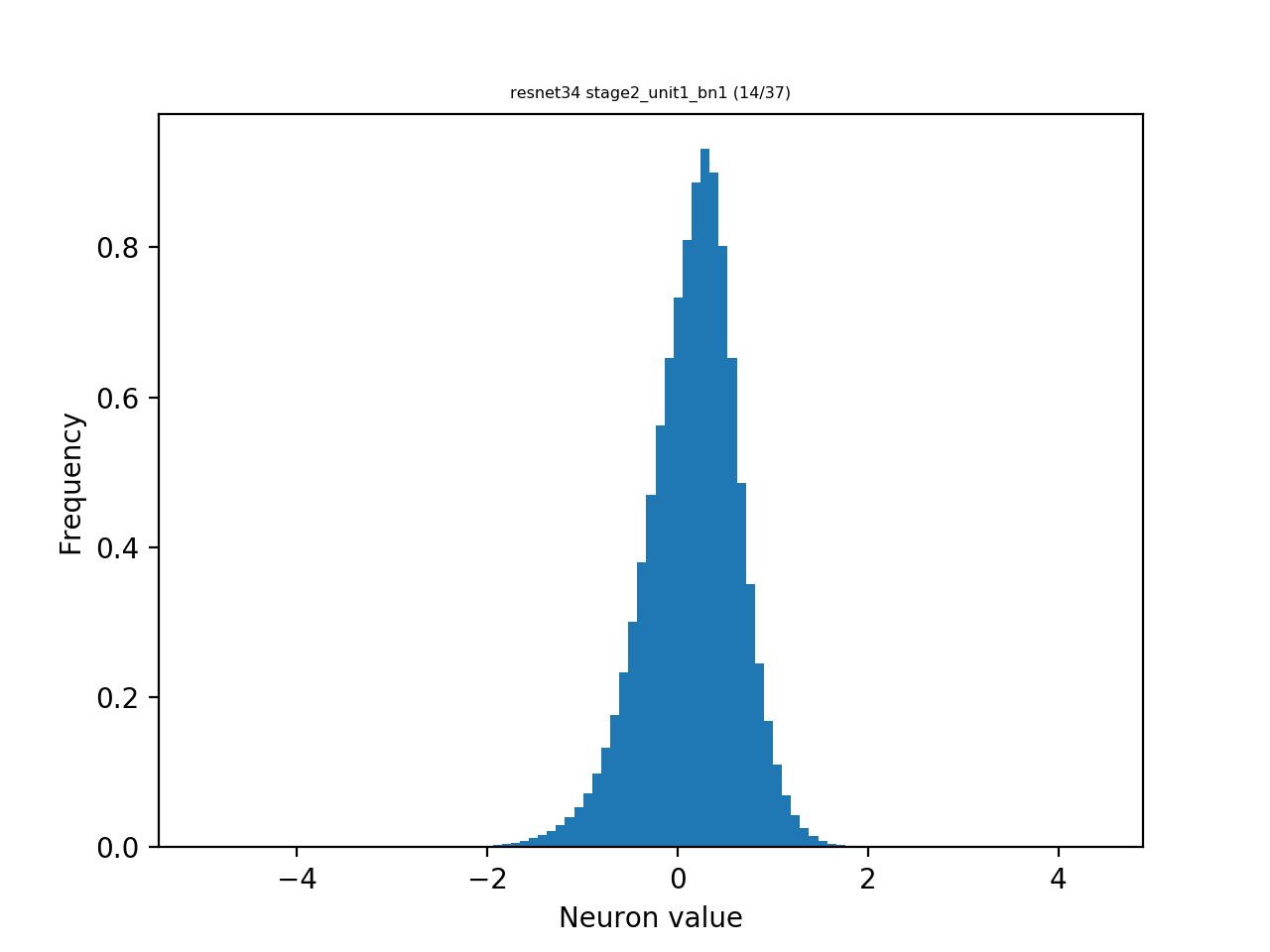}
    \caption{BatchNorm layer}
    \label{fig:batchnorm/histogram/batchnorm}
  \end{subfigure}%
  \begin{subfigure}{.5\textwidth}
    \centering
    \includegraphics[width=\linewidth]{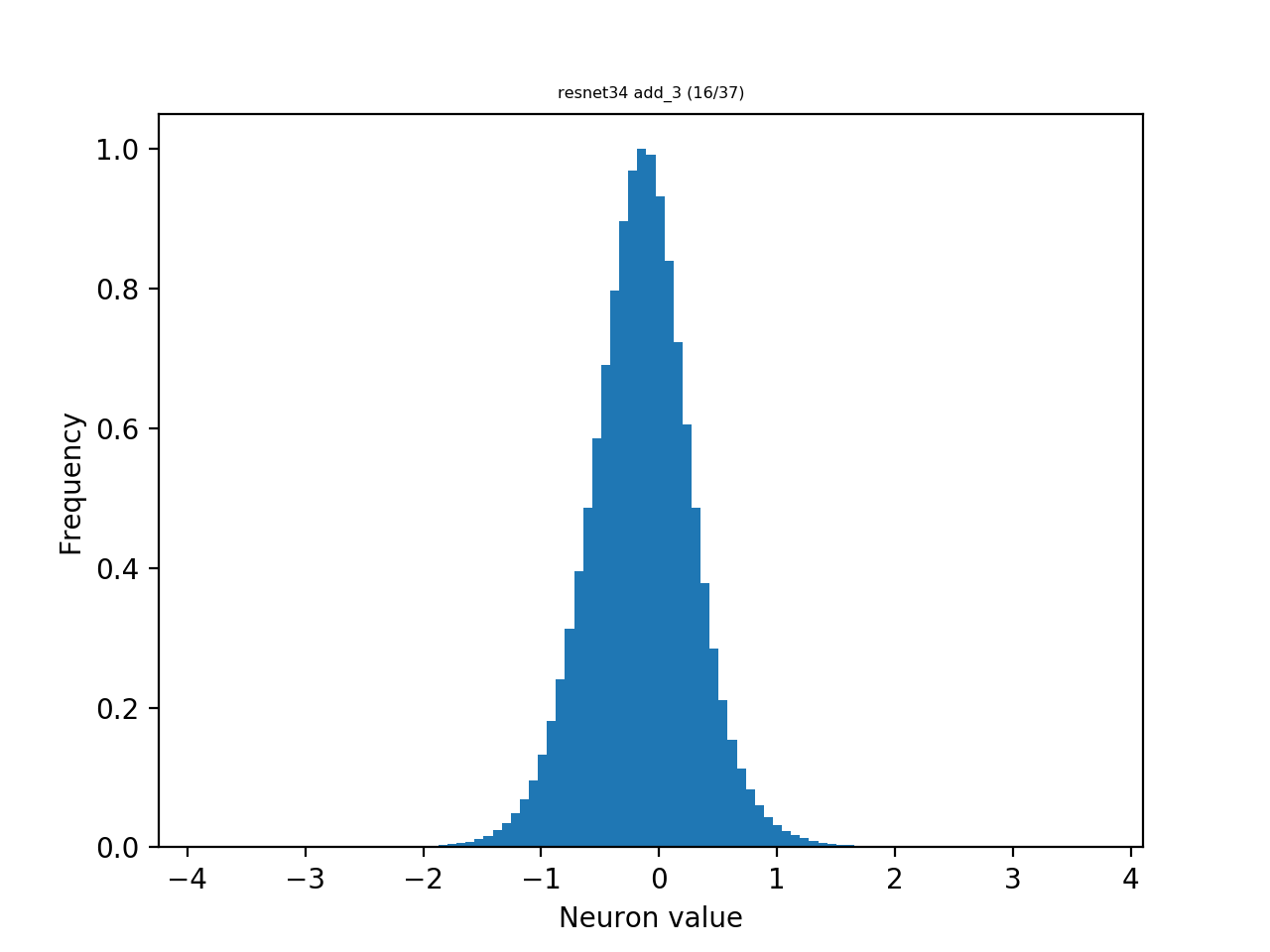}
    \caption{post-BatchNorm layer}
    \label{fig:batchnorm/histogram/post}
  \end{subfigure}
  \caption[BatchNorm histogram]{%
    Histograms of neuron output values at various layers.
    \csubref{fig:batchnorm/histogram/batchnorm} \texttt{stage2\_unit1\_bn1}
    (BatchNorm layer) of ResNet-34 with \protect$\mu=0.13, \sigma=0.50$.
    \csubref{fig:batchnorm/histogram/post} \texttt{add\_7} (post-BatchNorm
    layer) of ResNet-34 with \protect$\mu=-0.16, \sigma=0.47$.%
  }
  \label{fig:batchnorm/histogram}
\end{figure}

For ResNet models, the neuron output values of the BatchNorm layers
\emph{experimentally} appear to be approximately normally distributed with
their own mean and standard deviation.
\unresolved{
  Support with experimental evidence.
}
Thus, we assume that the random variable associated with a single neuron
output~$y_i$ is ${Y_i \sim \mathcal{N}(\mu_i, \sigma_i^2)}$. Additionally, the
mixture distribution of neuron output values (which we take to be the mean
average of the individual neuron output distributions) also
\emph{experimentally} appears to be roughly normally distributed: ${Y \sim
{\frac{1}{n} \sum_{i=1}^{n} \mathcal{N}(\mu_i, \sigma_i^2)} \approx
{\mathcal{N}(\mu, \sigma^2)}}$, where $n$ is the total number of neurons in the
layer. This is shown in \cref{fig:batchnorm/histogram/batchnorm} for a layer
from ResNet-34. Note that these are not mathematical facts and need not hold
for all models and layers.

Sometimes, the distribution of neuron output values for layers that follow
BatchNorm layers also experimentally appear to be normally distributed. This is
the case for the layer visualized in \cref{fig:batchnorm/histogram/post}. Note
that this isn't always the case: for example, a ReLU layer will clip negative
values in the left tail of the distribution, making the resulting distribution
non-normal. For a normal random variable, 99.7\% of values lie within 3
standard deviations from the mean, so the tensor values can be clipped to the
interval $[\mu - 3\sigma, \mu + 3\sigma]$ without significantly inaccurate
reconstruction of too many values.

\FloatBarrier

\subsection{Uniform quantization}
\label{sec:quantization/uniquant}

Uniform quantization involves constructing $N$ evenly sized bins $B = \{(x_0,
x_1], \allowbreak \dots, \allowbreak (x_{N-1}, x_N]\}$ over a finite interval
${[x_\textrm{min}, x_\textrm{max}]}$ and using a value between the bin edges as
the reconstruction value for that bin. The number of bins is also known as the
number of \emph{levels}~\cite{sayood2018}. In the following experiments, we
will use the midpoints of the bin edges as reconstruction values.
\resolved{
  Optimal uniform quantization involves using the \emph{centroids}
  (conditional means) of the bins as reconstruction levels. The centroid of a
  bin equals its midpoint only if the probability density in the bin is
  uniform; otherwise, it is moved towards the side where there is higher
  probability density.
}
\reply{
  I've adjusted the wording for now. I might later rerun the experiments with
  centroids derived using the experimentally determined distributions.
}

The process of quantization and dequantization is lossy; there is some error in
reconstruction. A good choice for the number of levels and clipping interval
will minimize the amount of distortion due to the imprecise reconstruction. The
specific choice is dependent on the output distribution at the given layer.

Consider the distribution for the \texttt{add\_3} layer of ResNet-34 shown in
\cref{fig:batchnorm/histogram/post}. Assuming normal distribution, 99.7\% of
possible neuron output values are contained within the interval ${[\mu - 3
\sigma, \mu + 3 \sigma]}$. Quantizing 32-bit float values over this interval to
256 levels (8-bits) gives us a compression ratio of 4:1 even without
variable-length coding. A drop of $< 0.5\%$ in top-1 accuracy is measured by
testing over 16,384 square-cropped images from the ILSVRC 2012 (ImageNet, 1000
classes) dataset~\cite{ILSVRC15}, resized to $224 \times 224$.

\begin{figure}
  \centering
  \begin{subfigure}{.5\textwidth}
    \centering
    \includegraphics[width=\linewidth]{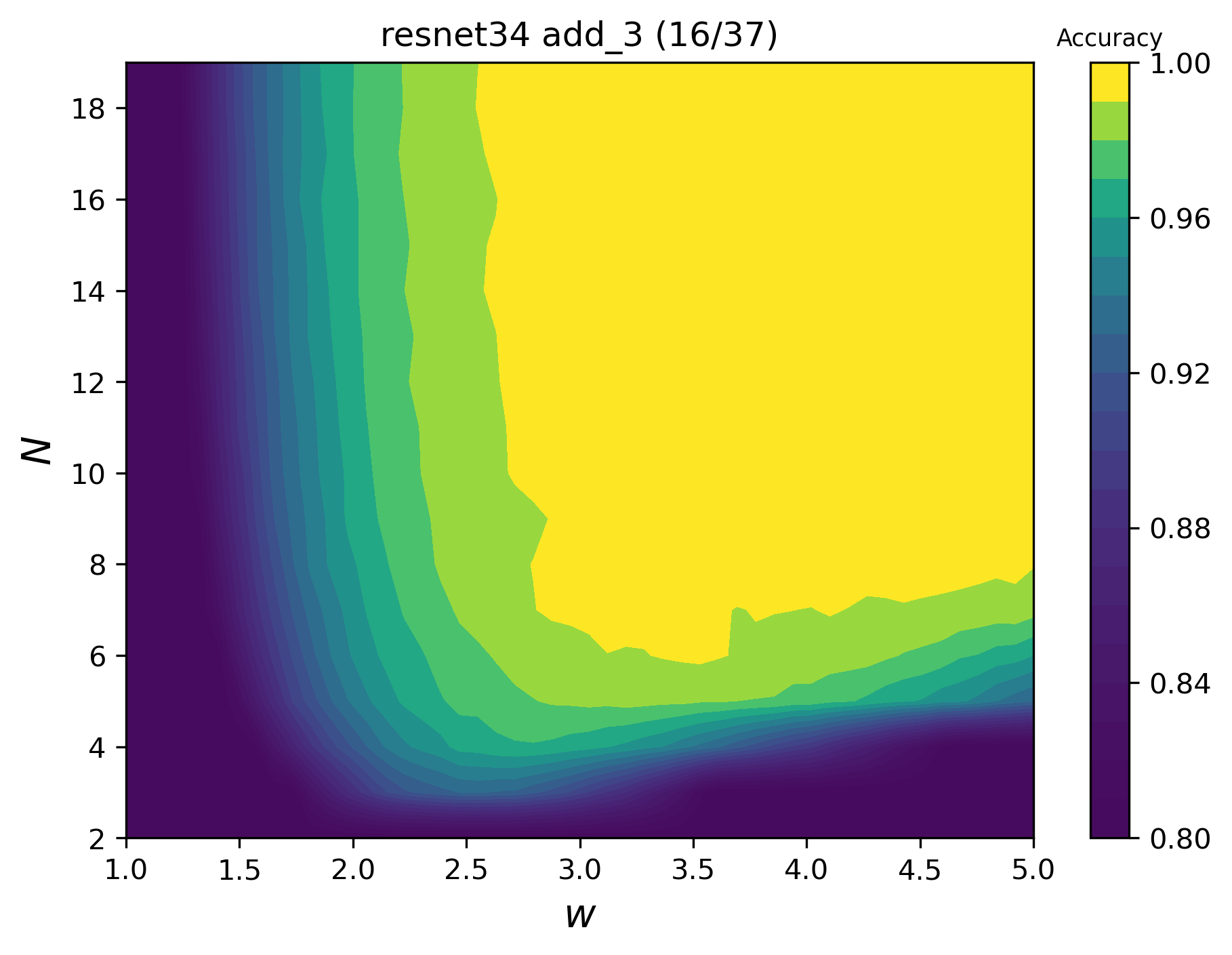}
    \caption{top-1 accuracy}
    \label{fig:quant/uniquant/accuracy}
  \end{subfigure}%
  \begin{subfigure}{.5\textwidth}
    \centering
    \includegraphics[width=\linewidth]{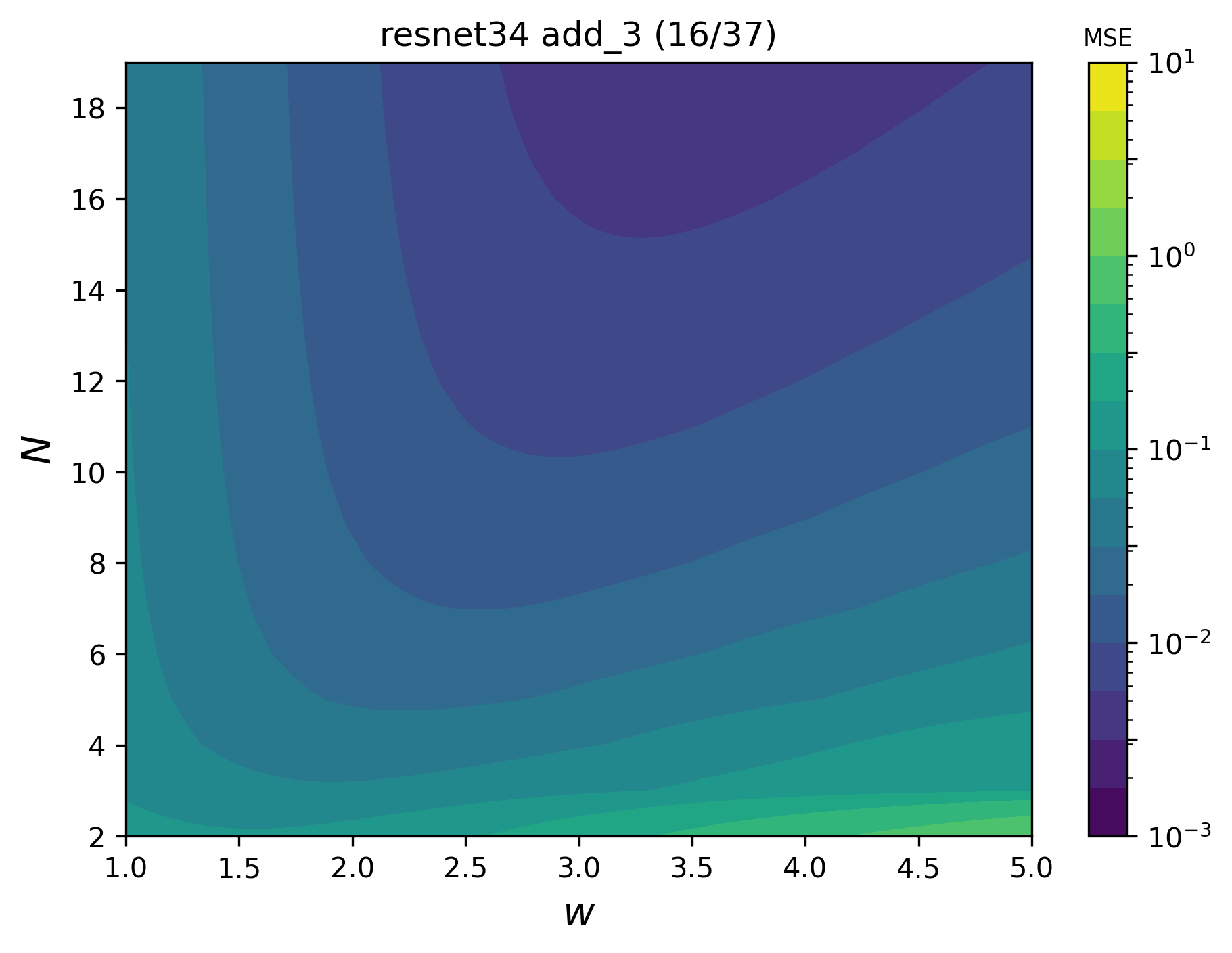}
    \caption{MSE}
    \label{fig:quant/uniquant/mse}
  \end{subfigure}
  \caption[Uniform quantization]{%
    Analysis of uniform quantization scheme with $N$ quantization levels and
    clipping range \protect$[\mu - w \sigma, \mu + w \sigma]$, where
    \protect$w$ is the clipping width (in units of \protect$\sigma$). %
    \csubref{fig:quant/uniquant/accuracy} Plot of inference accuracy of the
    reconstructed tensor resulting from quantization and dequantization of
    intermediate tensor. Accuracy is scaled relative to baseline accuracy
    computed from regular model inference on the given dataset. The darkest
    color represents \protect$\leq 80\%$ accuracy. %
    \csubref{fig:quant/uniquant/mse} Plot of MSE between the original tensor
    and the reconstructed tensor.
  }
  \label{fig:quant/uniquant}
\end{figure}

\begin{figure}
  \centering
  \begin{subfigure}{.5\textwidth}
    \includegraphics[width=\linewidth]{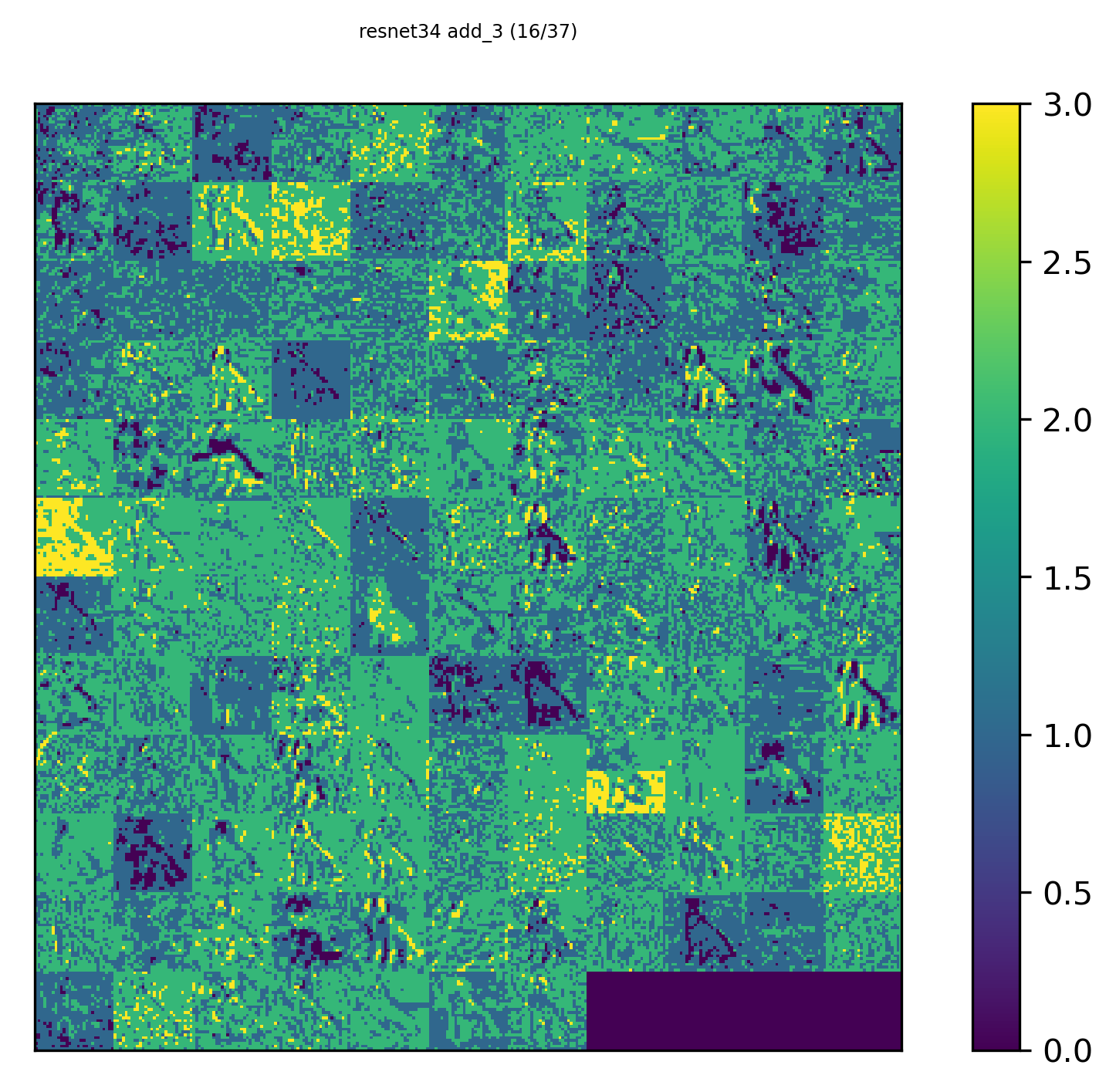}
    \caption{Uniform quantization}
    \label{fig:quant/featuremap/uniquant}
  \end{subfigure}%
  \begin{subfigure}{.5\textwidth}
    \includegraphics[width=\linewidth]{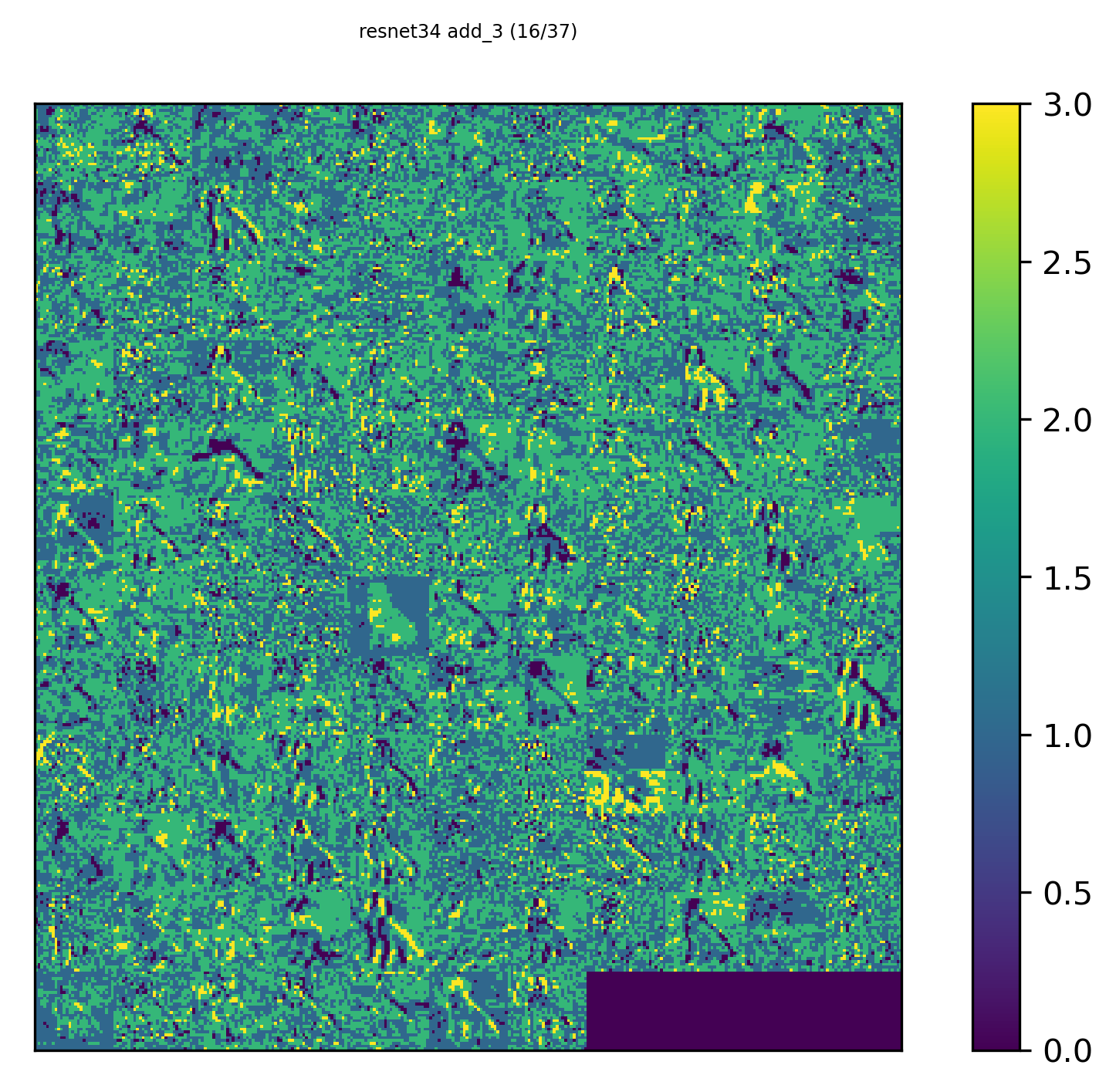}
    \caption{Uniform quantization (per-neuron distribution)}
    \label{fig:quant/featuremap/indepquant}
  \end{subfigure}
  \caption[Feature maps of quantized tensors]{%
    Feature maps of tensor data quantized by %
    \csubref{fig:quant/featuremap/uniquant} %
    uniform quantization with 4 levels over %
    \protect$[\mu - 3 \sigma, \mu + 3 \sigma]$ and %
    \csubref{fig:quant/featuremap/indepquant} %
    uniform quantization with 4 levels of neuron outputs normalized by their
    estimated distribution $Y_i$ over the interval \protect$[\mu_i - 3
    \sigma_i, \mu_i + 3 \sigma_i]$.%
  }
  \label{fig:quant/featuremap}
\end{figure}

We can further improve on this by imposing the constraint ${\mu = \frac{1}{2}
(x_\textrm{min} + x_\textrm{max})}$ (which fixes the center of the clipping
interval to the mean value). Define the clipping width $w$ by
${[x_\textrm{min}, x_\textrm{max}]} = {[\mu - w \sigma, \mu + w \sigma]}$.
Then, we can choose the clipping width and levels experimentally by plotting
the accuracy versus these two variables. This is done in
\cref{fig:quant/uniquant}, for 16,384 input samples from the same dataset as
above. Thus, we can achieve an accuracy drop of $< 0.5\%$ and a compression
ratio of 11:1 (without variable-length coding) by quantizing to 7 levels (2.8
bits/tensor element) over the interval ${[\mu - 3.3 \sigma, \mu + 3.3
\sigma]}$. Because the distribution of neuron output values is roughly
symmetric about the mean, it is unlikely that we can achieve significantly
better results by expanding the interval. The feature map of an example tensor
quantized using this method is shown in \cref{fig:quant/featuremap/uniquant}.

\FloatBarrier

\subsection{Uniform quantization (per-neuron distribution)}

\begin{figure}
  \centering
  \begin{subfigure}{.5\textwidth}
    \includegraphics[width=\linewidth]{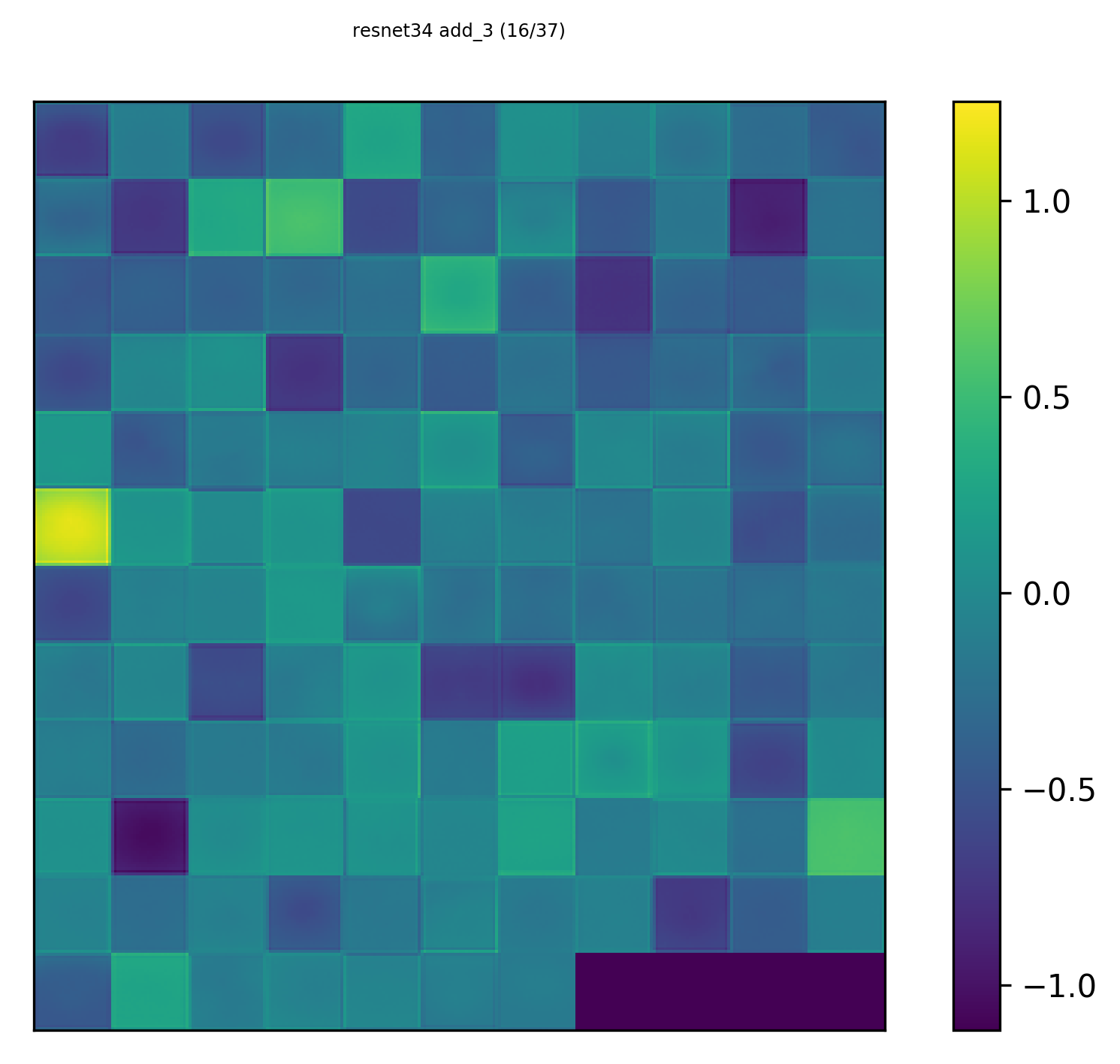}
    \caption{mean}
    \label{fig:quant/mean}
  \end{subfigure}%
  \begin{subfigure}{.5\textwidth}
    \includegraphics[width=\linewidth]{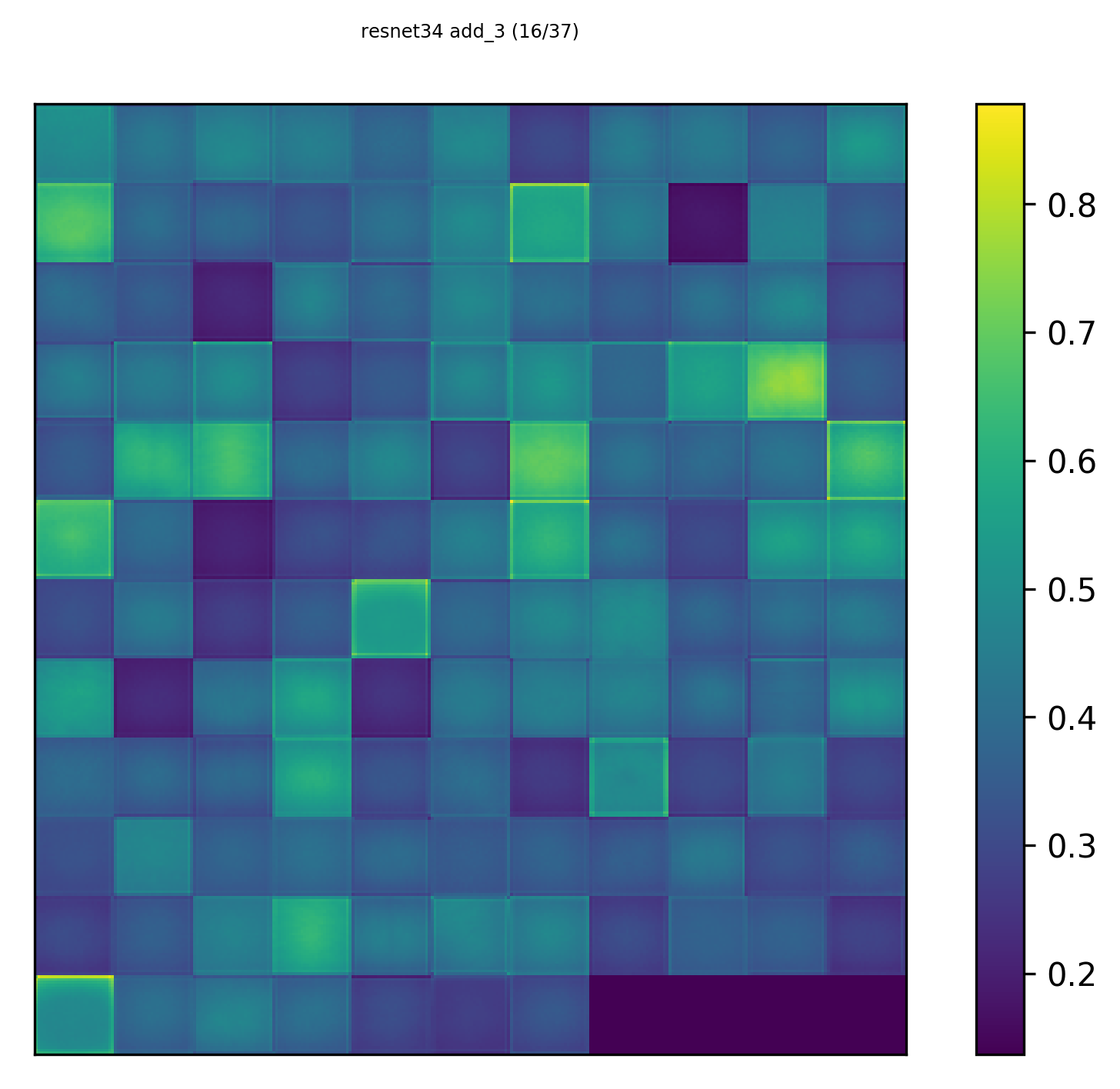}
    \caption{std}
    \label{fig:quant/std}
  \end{subfigure}
  \caption[Mean and standard deviation feature maps]{%
    The tensors resulting from 16384 different inputs were used to determine
    each neuron output's mean and standard deviation. The mean appears to be
    roughly the same for all neuron outputs within a given channel. The
    standard deviation is also roughly the same within each channel, though
    with some more variation.%
  }
  \label{fig:quant/mean_std}
\end{figure}

While treating a neuron output as a normal random variable from the aggregate
distribution shown in \cref{fig:batchnorm/histogram/post} provides a good set
of quantization bins, we can do better. Recognize that the output $y_i$ of a
single neuron has its own probability distribution. We may measure its
characteristics with the typical estimators of mean ${\mu_i = \mathbb{E}[Y_i]}$
and variance ${\sigma_i^2 = \mathbb{E}[(Y_i - \mu_i)^2]}$.
\cref{fig:quant/mean_std} shows the mean and standard deviations computed
individually for each neuron output, over 4096 sample images from the ILSVRC
2012 dataset.

Let ${U = (N, I)}$ be a uniform quantization strategy, where $N$ is the number
of levels and $I = [x_\textrm{min}, x_\textrm{max}]$ represents the clipping
interval. In \cref{sec:quantization/uniquant}, we were quantizing each neuron
output $y_i$ using a single uniform quantization strategy $U$ based on the
aggregate distribution associated with $Y$. We will now quantize each neuron
output $y_i$ individually by a uniform quantization strategy ${U_i = (N_i,
I_i)}$, based on the measured%
\footnote{In the typical, non-measure-theoretic sense of the word.} %
probability distribution associated with $Y_i$.

\begin{figure}
  \centering
  \begin{subfigure}{.5\textwidth}
    \centering
    \includegraphics[width=\linewidth]{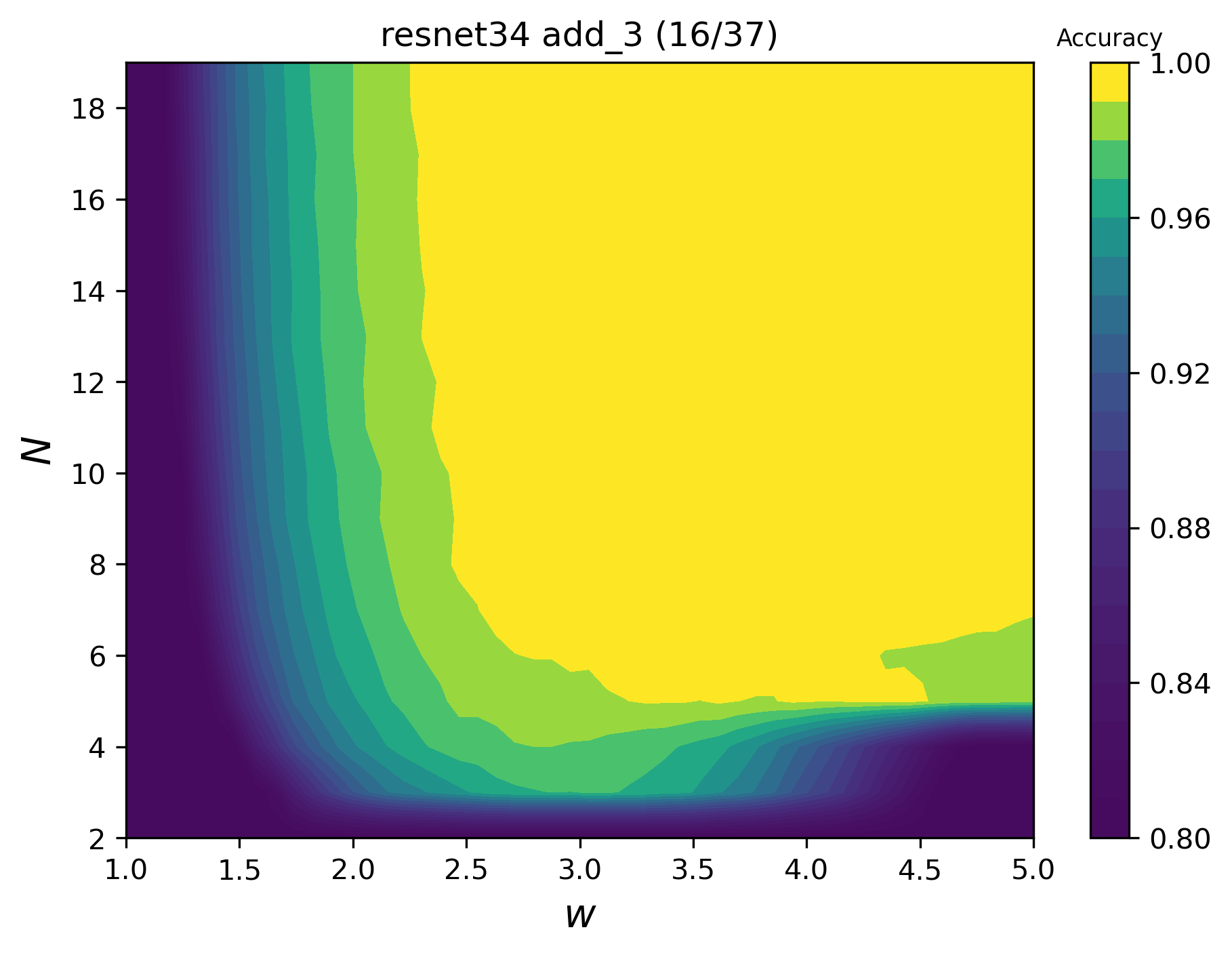}
    \caption{top-1 accuracy}
    \label{fig:quant/indepquant/accuracy}
  \end{subfigure}%
  \begin{subfigure}{.5\textwidth}
    \centering
    \includegraphics[width=\linewidth]{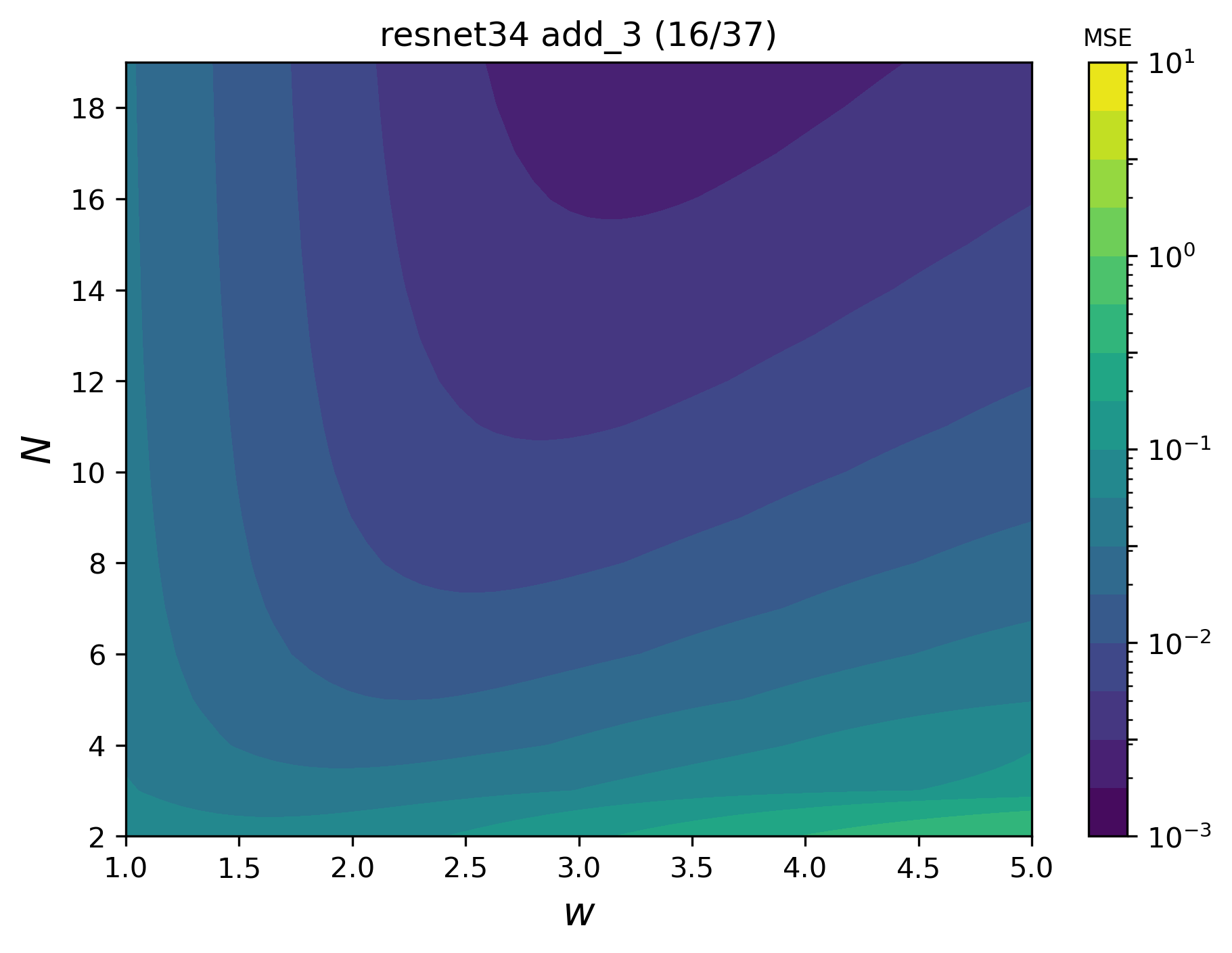}
    \caption{MSE}
    \label{fig:quant/indepquant/mse}
  \end{subfigure}
  \caption[Uniform quantization (per-neuron distribution)]{%
    Analysis of "per-neuron distribution" uniform quantization scheme with $N$
    quantization levels and clipping range \protect$[\mu_i - w \sigma_i, \mu_i
    + w \sigma_i]$, where \protect$w$ is the clipping width (in units of
    \protect$\sigma_i$), and \protect$\mu_i$ and \protect$\sigma_i$ are
    respectively the mean and standard deviation of the neuron output
    distribution \protect$Y_i$. %
    \csubref{fig:quant/indepquant/accuracy} Plot of inference accuracy of the
    reconstructed tensor resulting from quantization and dequantization of
    intermediate tensor. Accuracy is scaled relative to baseline accuracy
    computed from regular model inference on the given dataset. The darkest
    color represents \protect$\leq 80\%$ accuracy. %
    \csubref{fig:quant/indepquant/mse} Plot of MSE between the original tensor
    and the reconstructed tensor.%
  }
  \label{fig:quant/indepquant}
\end{figure}

Consider again the \texttt{add\_3} layer of ResNet-34. We impose the constraint
$\mu_i = {\frac{1}{2} ({x_\textrm{min}}_i + {x_\textrm{max}}_i)}$ and define
the clipping width $w_i$ by $I_i = [{x_\textrm{min}}_i, {x_\textrm{max}}_i] =
[\mu_i - w_i \sigma_i, \allowbreak \mu_i + w_i \sigma_i]$. To reduce the space
of possible quantization strategies, we fix the number of levels to a single
number, ${N_i = N}$. Furthermore, we also fix the clipping widths to a single
number, ${w_i = w}$. The feature map of an example tensor quantized using this
method is shown in \cref{fig:quant/featuremap/indepquant}. The accuracy of
inference on the reconstructed tensor and the MSE of reconstruction versus $N$
and $w$ are plotted in \cref{fig:quant/indepquant}. In comparison to
\cref{fig:quant/uniquant}, the MSE is lower. Furthermore, it takes fewer
quantization levels (and thus, bits) to obtain a $< 0.5\%$ drop in accuracy,
and a smaller clipping width. Thus, we can achieve an accuracy drop of $<0.5\%$
and a compression ratio of 12:1 (without variable-length coding) by quantizing
to 6 levels (2.6 bits/tensor element) over the interval $[\mu - 3.7 \sigma, \mu
+ 3.7 \sigma]$.

\FloatBarrier

\section{Reusing image codecs}

There has been much work already done in image compression. We can use this to
our advantage and attempt to repurpose image codecs to work with tensor data.
For this to occur, some additional work (e.g. tiling) needs to be done to
convert the tensor into the correct format for codec input.

\subsection{Tiling}
\label{sec:compression/tiling}

\begin{figure}
  \centering
  \includegraphics[width=\linewidth]{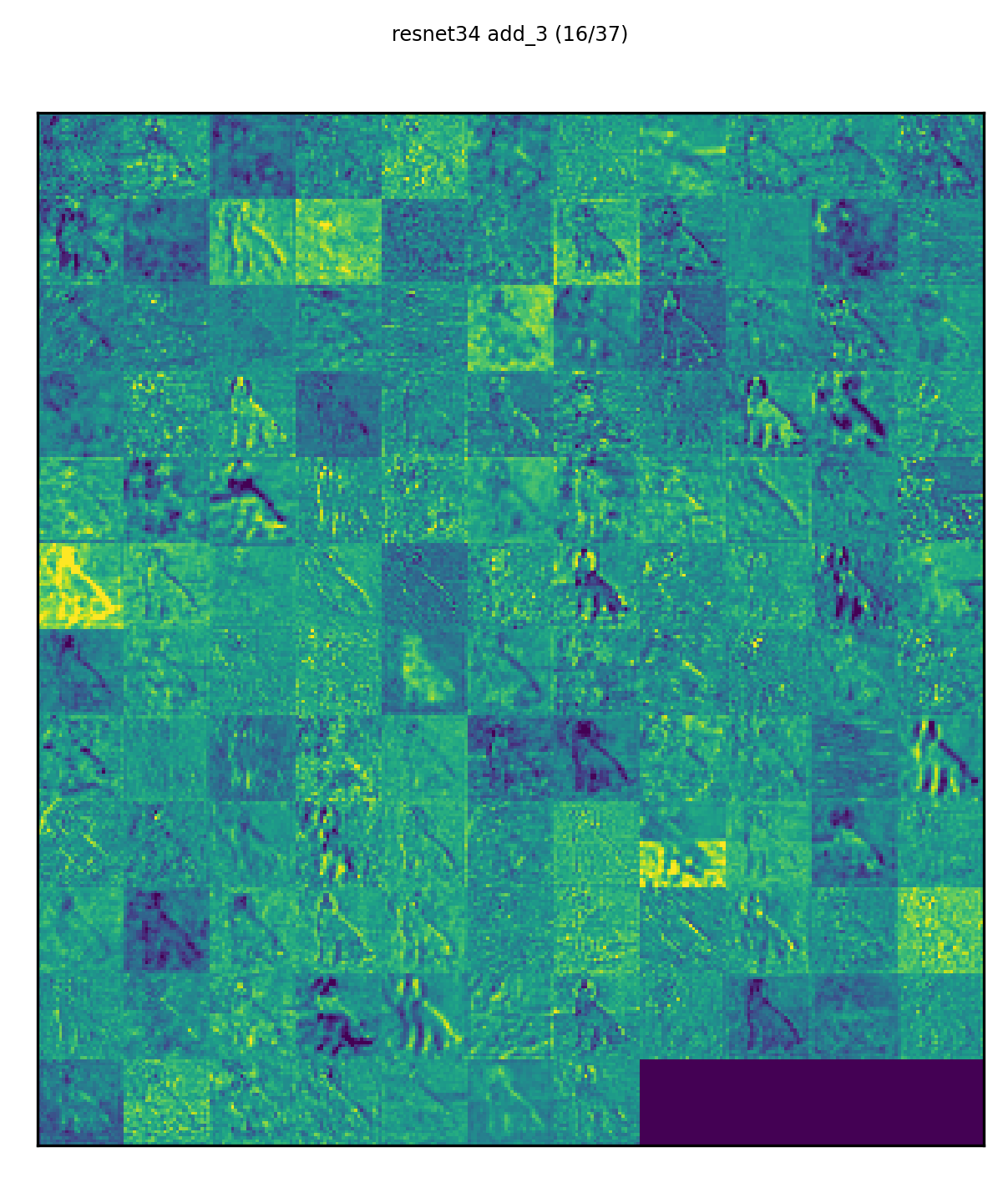}
  \caption[Feature map]{%
    Feature map of \texttt{add\_3} layer of ResNet-34 on input image of dog.
    Intra-channel redundancy exists in the form of similarities in intensity
    between nearby pixels. Inter-channel redundancies manifest themselves
    across clusters of channels. For instance, some channels have sharp edges
    outlining the dog; others and others contain few distinguishable edges. The
    similarities across channels are more apparent after normalizing the
    channels with respect to their mean \protect$\mu_c$ and variance
    \protect$\sigma_c$. However, these edges are not distinguishable in the
    feature maps of tensors quantized to a low level count, as in
    \cref{fig:quant/featuremap}. Nonetheless, such quantization only
    experiences \protect$< 0.5\%$ drop in accuracy, suggesting that the visual
    fidelity of the edges is not very important for performing accurate
    inference.%
  }
  \label{fig:featuremap}
\end{figure}

To allow a 3-d HWC tensor%
\footnote{%
  The indices of a HWC tensor are along the dimensions of height, width, and
  channels.%
}
to be inputted into an image codec, it must be reduced to a 2-d tensor, while
preserving useful structure (e.g. spatial locality) where possible. Consider
the feature map of the intermediate tensor of a CNN in \cref{fig:featuremap}.
The intermediate tensors of CNNs tend to contain many spatial redundancies
\emph{within} channels (and indeed, \emph{across} channels as well). We note
these redundancies in the following equations:
\begingroup
\addtolength{\jot}{1em}
\begin{align}
  T(y, x, c)
    \approx T(y + \Delta y, \, x + \Delta x, \, c)
    && \textrm{(intra-channel)}
    \\
  S(c, c')
    \geq d \left(
      \tfrac{1}{\sigma_c} [T(c) - \mu_c], \;
      \tfrac{1}{\sigma_{c'}} [T(c') - \mu_{c'}]
    \right)
    && \textrm{(inter-channel)}
  \label{eq:cnn_redundancies}
\end{align}
\endgroup
\noindent
where $T$ is a 3-d HWC tensor and $S : \{1, \ldots, C\} \times \{1, \ldots, C\}
\to \mathbb{R}$ represents some bound on the distance $d$ between a pair of
channels. Note that a conventional distance metric on vector spaces, such as an
$\ell_2$-norm, is not necessarily the best choice: the convolutional kernels
from the last convolutional layer may offset pixels, or invert the feature map.
It is perhaps better to apply the conventional distance measure on the
normalized channels after they have been "blurred" by a max-pool with the
stride length equal to the width of the convolution kernels. That is, ${d'(T_c,
T_{c'})} = {\Vert\textrm{maxpool}(T_c) - \textrm{maxpool}(T_{c'})\Vert_2}$.
Then, to account for channels with "negative" edge polarity (i.e. background
darker than the edges), one could set ${d(T_c, T_{c'})} = \min({d'(T_c,
T_{c'})}, \; {d'(T_c, -T_{c'})})$.


A tiling strategy that keeps nearby channels together involves simply assigning
each channel to a $C_W \times C_H$ grid, where each grid entry is of size $W
\times H$. The result should be a 2D image of size $(C_W \cdot W) \times (C_H
\cdot H)$. To ensure there are enough available grid entries for all the
channels, $C \leq C_W \cdot C_H$. One should also keep in mind that some image
codecs require or prefer that the image dimensions be reasonably similar.

Because the image will be storing a quantized tensor, it is preferable that the
color information be stored with an indexed color palette. However, not all
codecs support indexed or 8-bit color. For such codecs, the tensor value may be
stored in RGB or YUV color models. One should note that some codecs will
convert RGB to YUV prior to encoding (and back to RGB when decoding). This is
done because luminance (Y) and chroma (UV) can be encoded separately at
different rates; a common format is 4:2:0, which stores twice as many luminance
pixels as chroma pixels. If the entirety of the RGB color model is used, the
conversions to and from YUV can be destructive. Thus, when using RGB with such
codecs, it is best to represent a single tensor element as an RGB pixel with
the R, G, and B components sharing the same intensity value. When storing in
YUV directly, image codecs such as JPEG do not differentiate between
compression techniques and parameters used for Y and UV data, so it is
acceptable to store a tensor element in \emph{either} a Y, U, or V component.
In the case of JPEG, the Y, U, and V planes are compressed separately;
macroblocks only consist of one color component at a time. Some JPEG encoders
and decoders also support grayscale images which consist of only the Y
component.

\subsection{JPEG}

Though JPEG is a fairly old standard for image compression and many newer
codecs and technologies have been developed since its introduction in 1992, it
is still widely used, and it is a good starting point to consider given its
ubiquity. JPEG first transforms the input image by applying a discrete cosine
transform (DCT) to macroblocks ($8 \times 8$ or $16 \times 16$) to exploit
spatial redundancies, and then scaling by a cleverly weighted quantization
table%
\footnote{%
  Typically, the quantization table is precomputed. Most JPEG encoders such as
  libjpeg~\cite{libjpeg} use the IJG standard quantization
  tables~\cite{itut.t81.1992}. Another popular set of quantization tables are
  the proprietary Adobe Photoshop quantization tables. On the other hand, some
  encoders (typically those that are built into cameras) use adaptive
  quantization tables that are generated on-the-fly based on the image being
  compressed.~\cite{kornblum2008jpeg,kee2011digital}%
}
which allocates a larger number of bits to low frequency coefficients. The
resulting coefficients are then run-length encoded (RLE) and entropy coded via
Huffman coding.

%
%

By tiling the tensor in the manner described in \cref{sec:compression/tiling},
we are able to keep channel data contained within as few macroblocks as
possible and maintain contiguity across the height and width tensor dimensions.
This helps the DCT exploit spatial intra-channel redundancies. Unfortunately,
this comes at the cost of exploiting inter-channel redundancies: the DCT is
local to each macroblock, and thus cannot look across channels.

Images and video, the primary types of input data for CNNs, already possess
expert-designed compression codecs. Thus, in order to be competitive with a
server-only inference strategy (\cref{fig:strategies/server}), one needs to
compare the compressibility of input data with intermediate tensor data. Since
JPEG is lossy, in \cref{fig:accuracyvskb_jpeg} we compare the model accuracy
performance of applying JPEG compression to input data versus applying it to
tensor data at various output file sizes.

\providecommand{\fig}[4]{}
\renewcommand{\fig}[4]{%
  \begin{subfigure}[t]{.5\textwidth}
    \centering
    \includegraphics[width=\linewidth]
    {img/accuracyvskb/jpeg_uniquant256/resnet34/resnet34-#1of37-#2.png}
    \caption{ResNet-34, \ \texttt{#3} layer, \ #4}
    \label{fig:accuracyvskb_jpeg/resnet34_#2}
  \end{subfigure}%
}
\begin{figure}
  \centering
  \fig{08}{pooling0}{pooling0}{\hwc{56}{56}{64}}%
  \fig{16}{add_3}{add\_3}{\hwc{28}{28}{128}}
  \fig{22}{add_7}{add\_7}{\hwc{14}{14}{256}}%
  \fig{30}{add_13}{add\_13}{\hwc{7}{7}{512}}
  \caption[JPEG accuracy vs compressed output size]{%
    Top-1 image classification accuracy vs JPEG compressed frame output size
    across various layers of ResNet-34. For shared inference, tensor data was
    stored in the Y (luma) component in tiles. Pre-trained
    model~\cite{classification_models} trained on ImageNet (1000 classes).
    Tested on ImageNet. Results for more layers can be found in
    \cref{sec:appendix/additional_figures}.%
  }
  \label{fig:accuracyvskb_jpeg}
\end{figure}

The methodology for generating these plots is described as follows. The input
image dataset is generated by taking image samples from ImageNet (1000
classes), cropping them to a ratio of 1:1, downscaling to $224 \times 224$,
saving the resulting images again as JPEG images, and filtering out images that
are not within a file size of 30 $\pm$ 0.3 KB. From these, we keep a sample set
of 4096 images (with 4--100 images representing each class), and then compress
them via JPEG at a variety of quality levels for each image, generating over
100,000 compressed JPEG images at varying file sizes. The resulting JPEG images
are binned into logarithmically spaced intervals within the range 1 KB -- 30
KB, taking care that no bin contains more than one JPEG generated from the same
source image. Inference is then performed on these binned images, and the top-1
accuracy is computed for each bin. The result is plotted as the
\emph{server-only inference} curve.

The sample set of 4096 source images is then taken again, but this time the
images are run through the client-side model. The resulting tensors are then
quantized to 256 levels (8 bits) and tiled. Then, they are inserted into the
luminance channel and JPEG compression is applied to them at a variety of
quality levels, resulting in over 100,000 compressed tensors. The resulting
compressed tensor data is then binned using the same process as for the
server-only inference, ensuring that no bin contains more than one compressed
tensor generated from the same source image. These bins are then run through
server-side model inference and the resulting top-1 accuracies are plotted in
the \emph{shared inference} curve.

As expected, the model accuracy falls due to reconstruction errors as the
compressed output size shrinks. The point at which the accuracy falls below a
particular threshold varies depending on the split point. In general, the
accuracy curves tend to rise as we split at deeper layers in the model. Note
that the accuracy curves may still fall at deeper layers due to a variety of
reasons such as larger tensor dimensionality, changes in neuron output
distribution, and reduced resilience to reconstruction errors.

Because JPEG compression codecs such as libjpeg~\cite{libjpeg} and
libjpeg-turbo~\cite{libjpegturbo} offer no bitrate controls, it is tricky to
control the size of the output. The primary control for rate/distortion offered
by JPEG encoders is the \textbf{quality factor}, which ranges between 1-100 and
typically controls the DCT coefficient quantization table. Because of the
entropy coding which follows the quantization process, it is difficult to
predict the output size. This creates some difficulty in maintaining a constant
bitrate to adhere to network bandwidth and quality constraints and also in
maintaining a stable model accuracy. Indeed, if one is not careful, the bitrate
can fall into a range where inference accuracy is poor.

\subsection{JPEG 2000}

We now consider JPEG 2000~\cite{JPEG2000}, the successor to the JPEG format. In
contrast to JPEG's block-based DCT, JPEG 2000 uses a discrete wavelet transform
(DWT) which can be applied to the image globally. Thus, JPEG 2000 avoids
blocking artifacts and also exploits larger redundancies across the image.
Similar to JPEG, the resulting transform coefficients are quantized and entropy
coded, though with more sophisticated techniques. These techniques also enable
fine-grained bitrate control via rate-distortion optimization. This makes JPEG
2000 a more predictable codec than JPEG.


\cref{fig:accuracyvskb_jpeg2000} plots the model accuracy versus output size.
These are generated in a similar way to the JPEG curves in
\cref{fig:accuracyvskb_jpeg}; however, instead of trying all possible values of
an arbitrary control like quality level, JPEG 2000's bitrate controls are used
to precisely specify the desired output sizes. This is much quicker (less
samples need to be tried to find a matching output sizes), leads to a more
uniform distribution of samples, and reduces the variance in the dependent
variable (output size).

Surprisingly, for server-only inference, the resulting accuracy at low bitrates
is consistently worse than JPEG. This is somewhat unexpected, since JPEG 2000
should perform better than JPEG at low bitrates. This occurs possibly because
ImageNet images are already in the JPEG format and re-encoding JPEGs to JPEGs
while only changing the quality level parameter merely corresponds to a
rescaling of the quantization table; whereas transcoding from JPEG images to
JPEG 2000 images requires a more significant change in the compression
technology (from DCT to DWT) and must also deal with the JPEG artifacts present
in the source image. Another possible explanation is that the JPEG 2000 encoder
being used (OpenJPEG~\cite{openjpeg}) is not optimized for such low bitrates.

On the other hand, the shared inference curves sometimes provide useful
characteristics. For instance, to maintain an above 99\% accuracy in the
\texttt{add\_7} layer of ResNet-34, JPEG requires at least 9 KB tiled tensor,
whereas JPEG 2000 only requires 7 KB per tiled tensor. As bitrates get even
lower, JPEG once again begins to outperform JPEG 2000.

\providecommand{\fig}[4]{}
\renewcommand{\fig}[4]{%
  \begin{subfigure}[t]{.5\textwidth}
    \centering
    \includegraphics[width=\linewidth]
    {img/accuracyvskb/jpeg2000_uniquant256/resnet34/resnet34-#1of37-#2.png}
    \caption{ResNet-34, \ \texttt{#3} layer, \ #4}
    \label{fig:accuracyvskb_jpeg/resnet34_#2}
  \end{subfigure}%
}
\begin{figure}
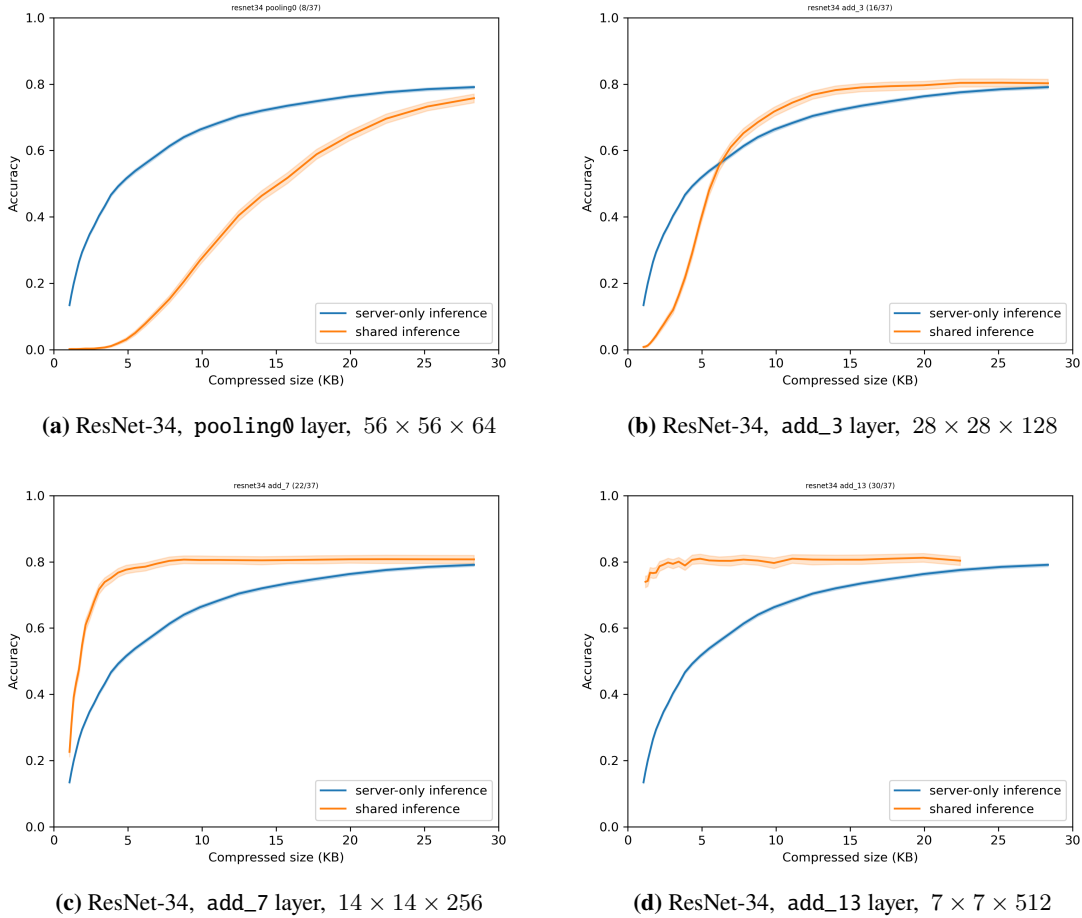

  \centering
  \fig{08}{pooling0}{pooling0}{\hwc{56}{56}{64}}%
  \fig{16}{add_3}{add\_3}{\hwc{28}{28}{128}}
  \fig{22}{add_7}{add\_7}{\hwc{14}{14}{256}}%
  \fig{30}{add_13}{add\_13}{\hwc{7}{7}{512}}
  \caption[JPEG 2000 accuracy vs compressed output size]{%
    Top-1 image classification accuracy vs JPEG 2000 compressed frame output
    size across various layers of ResNet-34. Results for more layers can be
    found in \cref{sec:appendix/additional_figures}.%
  }
  \label{fig:accuracyvskb_jpeg2000}
\end{figure}

\section{Summary}

In this chapter, we saw that models containing BatchNorm layers provide good
candidate layers to split upon. The aggregate output distributions of BatchNorm
layers and the layers that follow BatchNorm layers often approximate normal
distributions or other light-tailed distributions. Furthermore, the individual
neuron outputs of these layers can be treated as light-tailed distributions
over a large number of input samples. We also saw that the means of neuron
outputs, computed over a large number of samples, were reasonably uniform
across a given channel. Their standard deviations also showed spatial
uniformity in the x and y directions for a fixed channel.

Next, we looked at how applying uniform quantization with various intervals and
levels affected final inference accuracy. For the ResNet-34 model at the
\texttt{add\_3} layer, we saw a drop in top-1 accuracy by $< 0.5\%$ when
quantizing to 7 levels (2.8 bits/tensor element). Normalizing the tensor by the
by computed means and standard deviations for each individual neuron output
before quantizing allowed us to reduce the level count to 6 levels (2.6
bits/tensor element) in order to achieve the same top-1 accuracy.

Next, we looked at how to bring existing image codecs to work with tensor data.
To allow this, we first reshaped the 3-d tensor data into a 2-d format. Because
image codecs rely upon spatial redundancies within an image to compress the
image, it is important to reshape in such a way that tensor elements with
similar intensity values are kept close together. For CNNs, we noted two
primary redundancies: intra-channel and inter-channel. To exploit the
intra-channel redundancies, we tiled the channels into a 2-d grid. Next, we
quantized the reshaped data to 8-bit integers and then fed it into the JPEG and
JPEG 2000 codecs. Increasing the amount of image compression we applied to each
tensor reduced the top-1 accuracies. We discovered that as we progress through
the model's layers, the top-1 accuracy for a given compression ratio
\emph{usually} increases. This is due to reduced dimensionality of the tensor,
changes in the neuron output distributions, and increased resilience to
reconstruction errors. At a certain layer, shared inference accuracy becomes
larger than server-only inference accuracy for some size of compressed data.
For ResNet-34 with JPEG compression, shared inference consistently met or
outperformed server-only inference across all compression sizes at the
\texttt{add\_7} layer, which takes roughly a third of the total computational
expense of the model to compute.

As we will see in \cref{sec:implementation/latency_model}, this provides
promising results, and demonstrates how shared inference may situationally
outperform server-only inference for some combinations of hardware and network
conditions. With better compression methods that can exploit the inter-channel
redundancies in addition to the intra-channel redundancies we made use of here,
it is likely that shared inference can outperform server-only inference at
perhaps even earlier layers, further improving the usefulness of the shared
inference strategy.

\chapter{Towards tensor streams}
\label{chap:tensor_streams}

In the previous chapter, we looked at how to compress single tensors for
transmission and inference. But often, one is concerned with the inference of a
\emph{sequence} of inputs. For video, there is typically a strong relationship
between successive input frames. In this chapter, we will show that the
sequence of intermediate tensors generated from the input sequence also share
temporal similarities. Furthermore, during transmission of a real-time tensor
stream, some tensor data may be lost or corrupted; we will also discuss how to
continue with inference despite the loss of this information.

\section{Input transformations}
\label{sec:tensor_streams/input_transformations}

In order to develop techniques for encoding a stream of tensors, we should use
typical transformations of the input to determine what changes in the
intermediate tensor are typical.

For video, common transformations that occur within the input frame include
translation (panning/tilting), rotation, and rescaling (zoom) of groups of
pixels. As such, one of the key ideas used by video codecs is the idea of
motion compensation~\cite{woods2012}. This involves estimating the motion
vectors for the current frame with respect to a reference frame. A prediction
of the current frame is reconstructed by mapping the motion vectors over the
reference frame. The difference is then computed between the current frame and
the predicted frame; typically, this difference is small in magnitude. Along
with the motion vectors, the difference is then encoded and stored. Due to the
temporal redundancies in a sequence of frames, this can often result in
significant savings in comparison to encoding the current frame without motion
compensation.

\begin{figure}
  \centering
  \includegraphics[width=\textwidth]{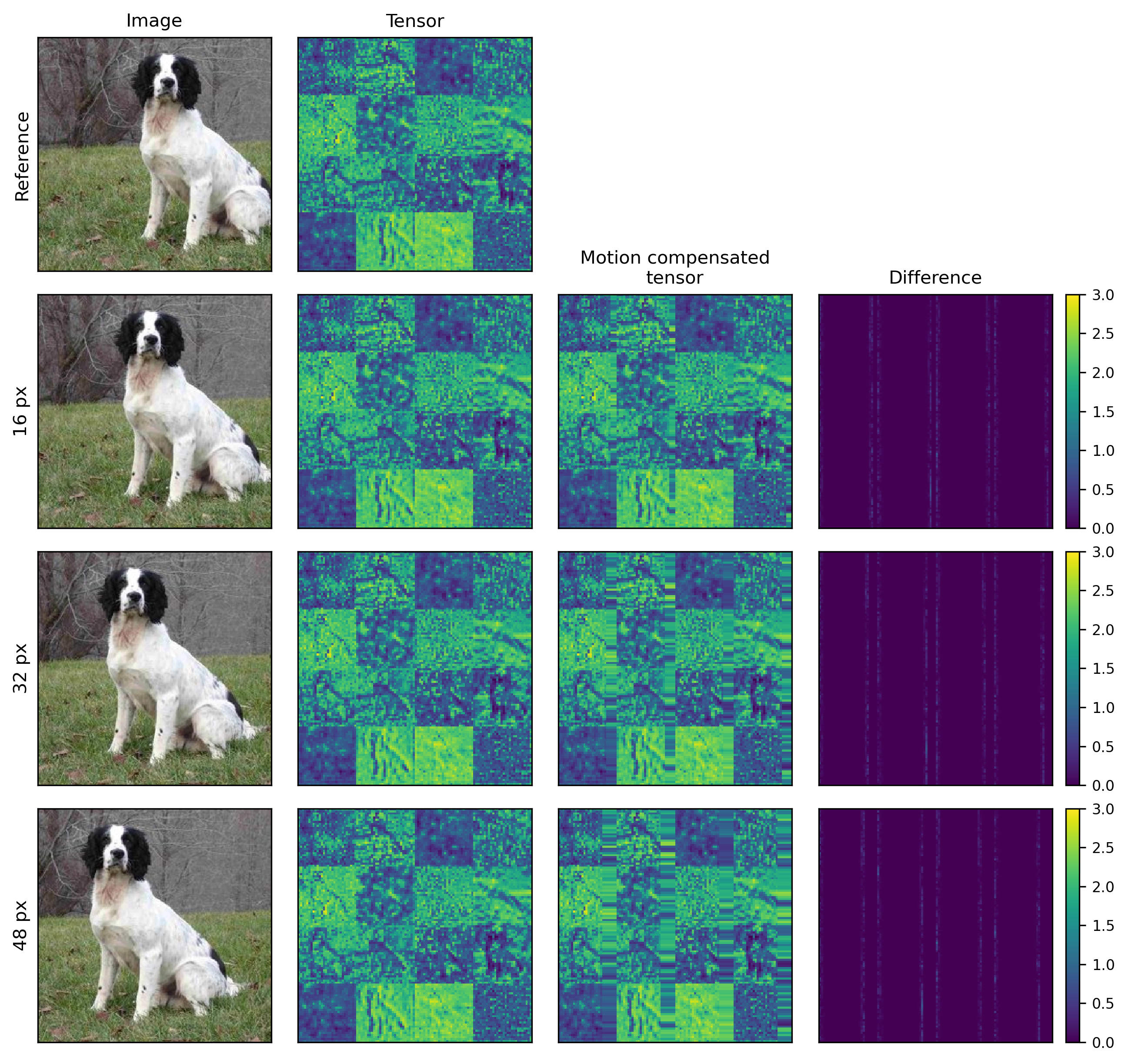}
  \caption[Input translated]{%
    The input image (\protect$224 \times 224 \times 3$) is horizontally
    translated by 16px, 32px, and 48px with respect to the initial reference
    frame (top). This corresponds to estimated horizontal translations in the
    intermediate tensor of the \texttt{add\_3} layer of ResNet-34 (\protect$28
    \times 28 \times 128$) by 2px, 4px, and 6px, respectively. The second
    column shows a select portion of the feature map of the intermediate tensor
    corresponding to each translated image. The third column shows the feature
    map of a tensor predicted by applying motion compensation on the reference
    tensor. The fourth column shows the absolute difference between the ground
    truth and the prediction, where the boundary regions that cannot be motion
    compensated have been zeroed. The PSNR of the non-boundary regions ranges
    between 90--92 dB. Interestingly, the difference in the interior regions of
    each channel is exactly 0. This occurs due to a combination of the
    translation invariance of preceding convolutional layers and strides of
    preceding pooling layers. Note that the colormaps have been adjusted for
    visual clarity.%
  }
  \label{fig:input/translate/pixel}
\end{figure}

Here, we consider global translation of frames. For simplicity, we will only
consider inputs which are translated horizontally and equally for every pixel.
In \cref{fig:input/translate/pixel}, we translate the reference frame by
amounts that are integer multiples $k \frac{w_I}{w}$ of the image width $w_I$
divided by the tensor width $w$. In a CNN such as ResNet-34, this induces,
roughly, a translation of the tensor $T$ in the $x$ direction by $k$ pixels.
This is due to the translation invariance, which comes from the fact that all
preceding layers are convolutional, pooling, batch normalization, ReLU, or
addition layers from the skip connections.

We can form a reconstruction $\hat{T}$ of the original tensor by:
\begin{equation}
  \hat{T}(y, x, c) = T_\textrm{ref}(y + v_y(x, y), \, x + v_x(x, y), \, c)
\end{equation}
where $v : \mathbb{R}^2 \to \mathbb{R}^2$ is computed through motion estimation
of the current input frame with respect to reference input frame, and rescaling
the resulting vectors by $\frac{w_I}{w}$. To assess the correctness of the
reconstruction, we define the peak signal-to-noise ratio (PSNR) for a tensor
$T$ and its reconstructed tensor $\hat{T}$ by:
\begin{gather}
  \textrm{MSE}
    = \lVert \hat{T} - T \rVert_2^{\; 2}
    = \frac{1}{HWC} \sum_{x=1}^{W} \sum_{y=1}^{H} \sum_{c=1}^{C}
    \left( \hat{T}(y, x, c) - T(y, x, c) \right)^2
    \\
  R
    = \max T - \min T
    \\
  \textrm{PSNR}
    = 10 \log \frac{R^2}{\textrm{MSE}}
\end{gather}

However, since not every pixel has a valid motion vector w.r.t. the reference
frame, we will ignore those regions in our calculations, for this synthesized
example. For these translated frames, the PSNR for the valid regions is
computed to be 90--92 dB. In fact, the interior regions of the reconstructed
tensor are \emph{exactly} equal to the original tensor. This is likely due to
the alignment of the pooling layers.

\begin{figure}
  \centering
  \includegraphics[width=\textwidth]{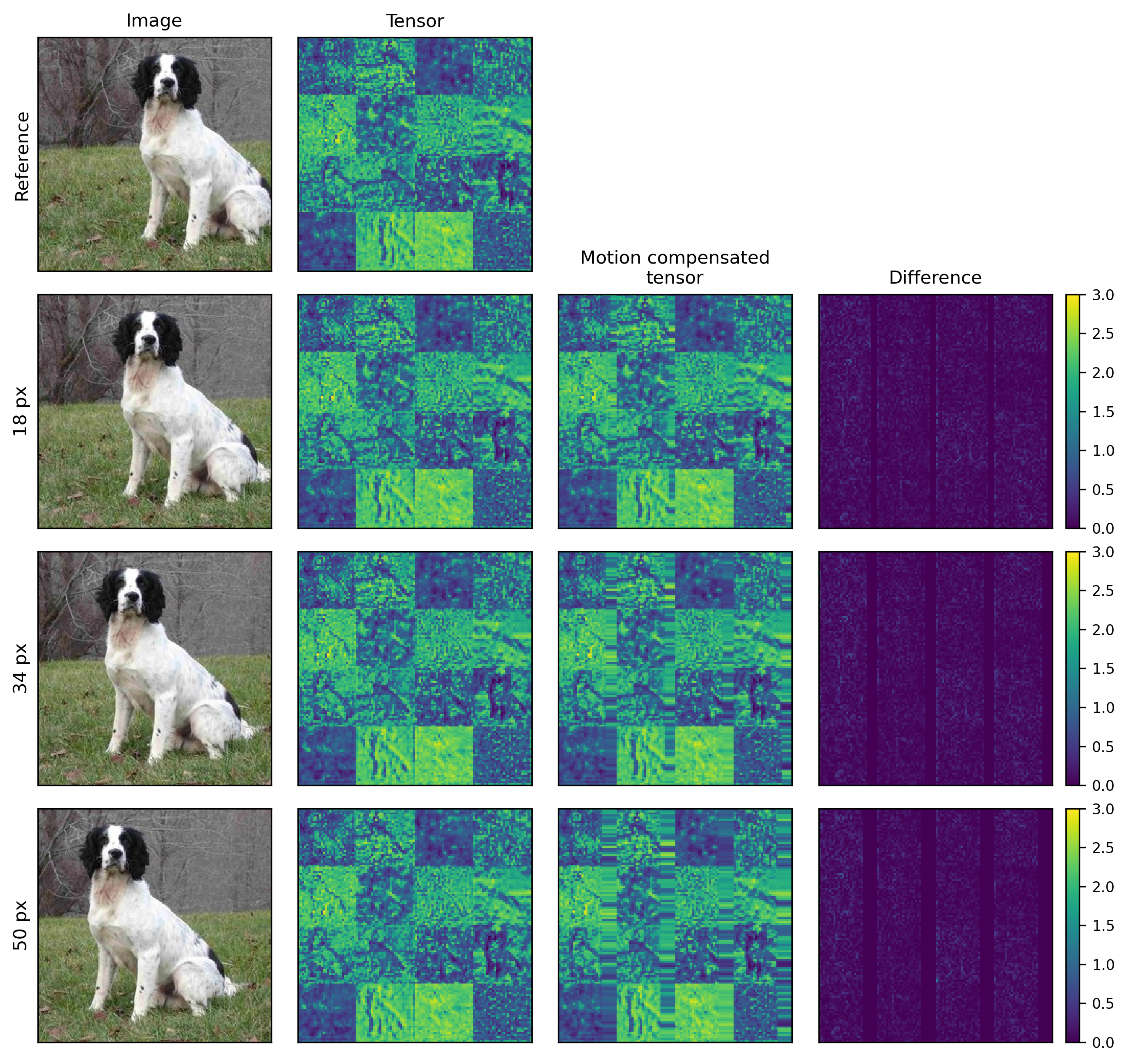}
  \caption[Input translated (subpixel)]{%
    The input image (\protect$224 \times 224 \times 3$) is horizontally
    translated by 18px, 34px, and 50px with respect to the initial reference
    frame (top). This corresponds to estimated horizontal translations in the
    intermediate tensor of the \texttt{add\_3} layer of ResNet-34 (\protect$28
    \times 28 \times 128$) by 2.25px, 4.25px, and 6.25px, respectively. The
    second column shows a select portion of the feature map of the intermediate
    tensor corresponding to each translated image. The third column shows the
    feature map of a tensor predicted by applying motion compensation on the
    reference tensor. The fourth column shows the absolute difference between
    the ground truth and the prediction, where the boundary regions that cannot
    be motion compensated have been zeroed. The PSNR of the non-boundary
    regions ranges between 74--75 dB. Note that the colormaps have been
    adjusted for visual clarity.%
  }
  \label{fig:input/translate/subpixel}
\end{figure}

To get a more realistic picture for the effect of global translations in
reconstruction, we let $k$ be non-integer so that the motion estimated vectors
for the tensor are also non-integer. \cref{fig:input/translate/subpixel} shows
an example of this. Unsurprisingly, the reconstruction error has increased,
resulting in a PSNR of 74--75 dB.

Motivated by these observations, one can imagine the following compression
strategy: precompute the motion compensation vectors on a scaled version of the
input frame. (This should be easier to do on the input frame rather than a
given channel since the input is less noisy.) Then, reuse the computed motion
vectors across all channels. Such an approach is both more computationally
efficient and potentially allows for better compression since different sets of
motion vectors do not have to be stored for each channel individually.

\FloatBarrier

\section{Error concealment}
\label{sec:tensor_streams/error_concealment}

In the event that some tensor data is corrupted or lost during transmission, we
should still attempt to give the best inference possible with the data that
remains. In this section, we investigate common forms of data loss and seek to
determine a tensor reconstruction strategy in the event of data loss. More
advanced strategies for recovering data utilize tensor completion
methods~\cite{Bragilevsky2020}, but we will not be considering those here. It
should be emphasized that the primary objective here is not to minimize
reconstruction error; rather, it is to obtain the best inference accuracy.

\begin{figure}
  \centering
  \begin{subfigure}{.5\textwidth}
    \centering
    \includegraphics[width=\linewidth]{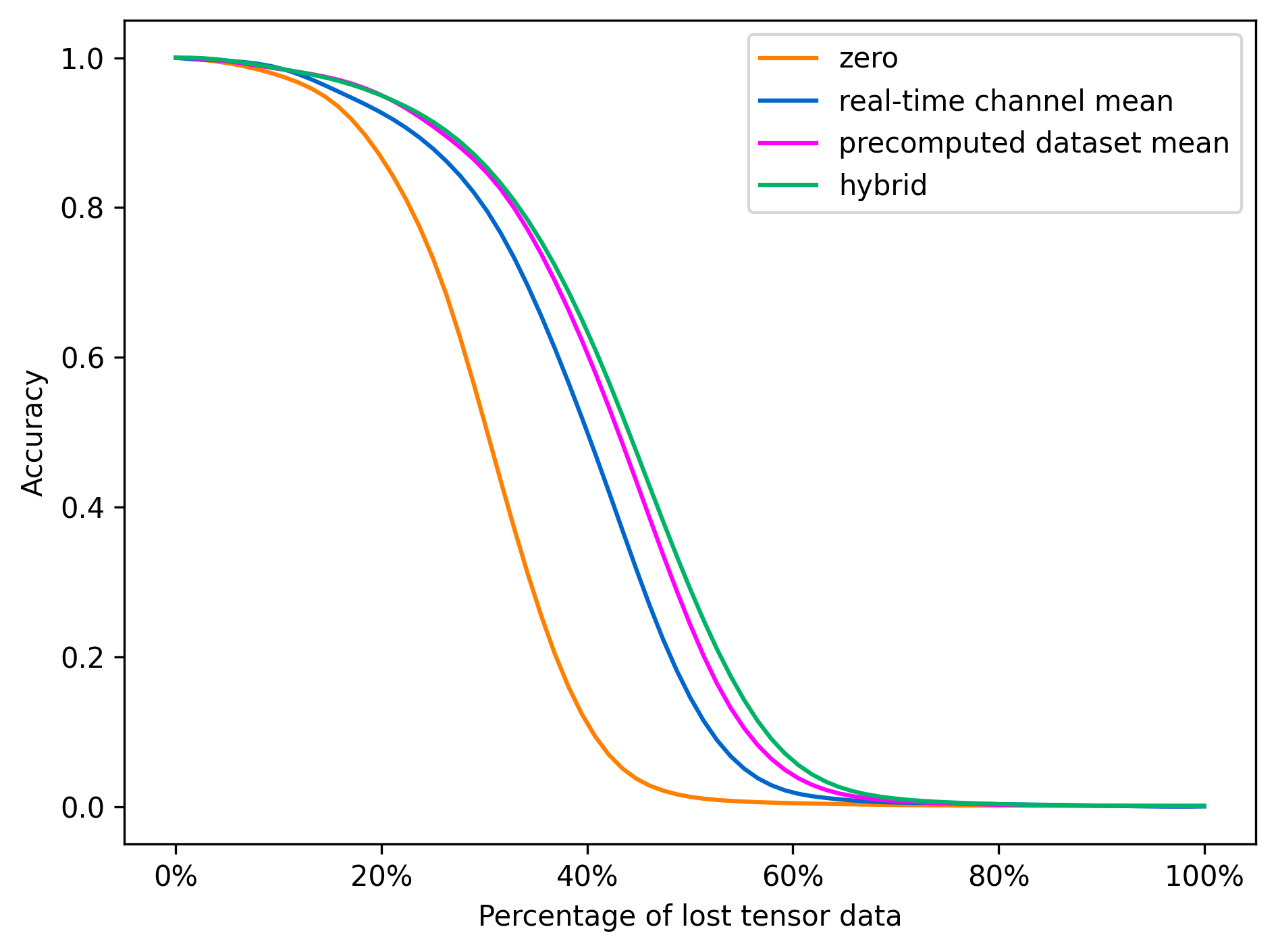}
    \caption{data loss by neuron}
    \label{fig:experiments/dataloss/neuron}
  \end{subfigure}%
  \begin{subfigure}{.5\textwidth}
    \centering
    \includegraphics[width=\linewidth]{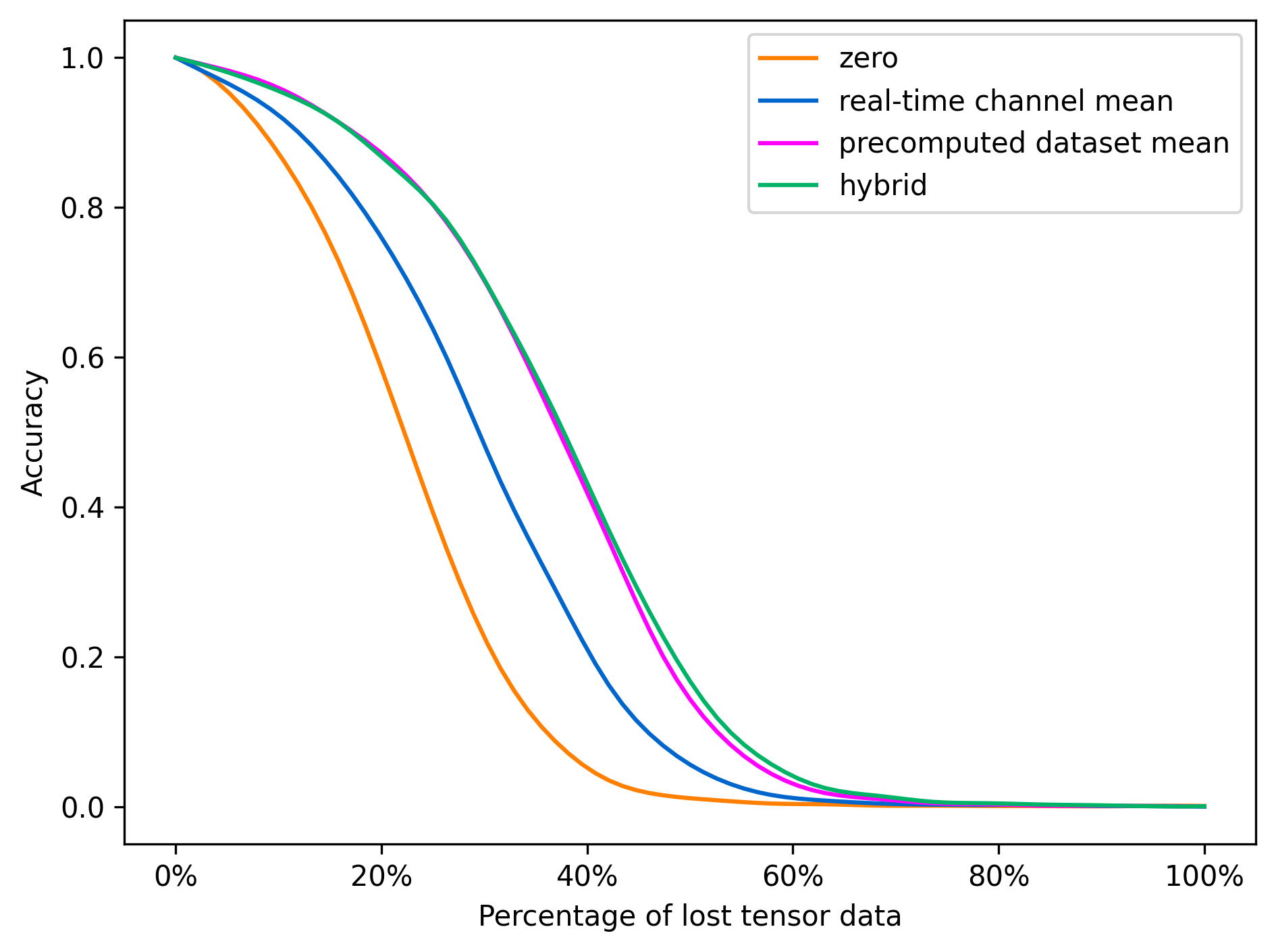}
    \caption{data loss by channel}
    \label{fig:experiments/dataloss/channel}
  \end{subfigure}
  \caption[Data loss accuracy]{%
    Accuracy versus amount of data loss, with application of various data
    recovery strategies on \texttt{add\_3} layer of ResNet-34.%
  }
  \label{fig:experiments/dataloss}
\end{figure}

\cref{fig:experiments/dataloss/neuron} summarizes the performance of various
data recovery strategies for the data loss of random tensor elements from a
particular layer of ResNet-34, and \cref{fig:experiments/dataloss/channel}
summarizes analogous results for the data loss of entire tensor channels. Each
data point is computed over 16,384 samples representing 1000 classes%
\footnote{%
  Each class is represented by 4--100 samples.%
}
from the ILSVRC 2012 dataset. The various strategies are described in order
below.

The simplest strategy for dealing with missing tensor data is to fill that
portion of the tensor with the value 0. As expected, this is the poorest
strategy and produces a curve that drops fairly quickly in terms of inference
accuracy (and would likely perform even worse for layers that do not generate
tensor data with near-zero mean).

A second strategy is to set the missing tensor data to the mean of its channel.
This requires the client to compute the channel-by-channel means and transmit
these means alongside the remaining tensor data. When $C \ll H W C$, the amount
of additional bandwidth taken by these means is minimal, though prediction,
quantization, and entropy coding techniques can likely be applied to them as
well if needed.

A third strategy that fares better is to set the missing tensor data to the
statistical mean of tensor values derived over some set of known data samples.
In this case, the samples are taken from partial inferences of ImageNet (1000
classes); in fact, a feature map of this particular statistical mean of tensor
values was shown earlier in \cref{fig:quant/mean}. The performance of this
strategy should improve further if the mean is taken over a more representative
sample set --- for instance, sample tensors generated from the previous frames
of a video, the audio background noise, or the previous audio chunks of an
audio stream. In contrast to the second strategy, this does not require any
additional data transmission.

A fourth strategy involves taking a hybrid approach to previous two strategies:
the reconstructed missing tensor element $\hat{T}$ is defined by $\hat{T}(y, x,
c) = \mu(y, x, c) + \left( T_\mu(c) - \frac{1}{H W} \sum_{i, j} \mu(i, j, c)
\right)$, where $\mu$ is the mean of tensors over a representative sample set,
and $T_\mu$ is the transmitted channel-by-channel mean.

These strategies perform roughly the same regardless of whether it is random
tensor elements or entire channels that are missing. Overall, filling in
missing tensor elements with their expected value precomputed over the dataset
seems to perform better than setting it to the channel's actual mean computed
at runtime. The hybrid approach of these does not perform any better at low
proportions of missing tensor data, and only marginally performs better at
higher proportions.

These methods do not make use of the various spatial and temporal redundancies
that occur within a tensor stream. Regardless, they perform well enough for low
proportions of missing tensor data: if 5\% of the tensor is missing, there is
only a 0.2\% drop in accuracy; for 10\% missing, a 2\% drop in accuracy; and
for 20\% missing, a 5\% drop. If the drop in accuracy meets an acceptable
threshold, these methods can provide a simple and computationally inexpensive
way of recovering the tensor. If a sufficiently large portion of the tensor is
lost, one can then swap to more sophisticated methods of tensor recovery, in
order to maintain an accuracy threshold.

\FloatBarrier

\section{Summary}

In this chapter, we looked at some useful properties within the tensor stream
that can be exploited for better tensor stream compression. For CNNs acting on
video input, successive tensors within the stream respond in the same way as
successive video frames do to common transformations (e.g. translations) within
the image. Furthermore, errors in a tensor stream can be partly concealed by
reasonably simple methods such as filling in missing tensor elements with the
mean value. This is because the end goal is the accuracy of inference rather
than merely reducing reconstruction error. Simple error concealment methods
such as these are also useful because of their predictability and lower
computational resource load in comparison to more advanced techniques.

\chapter{Implementation}
\label{chap:implementation}

Though the primary challenge for collaborative intelligence is compression,
there are also other challenges that occur in practice. Such challenges include
the work needed to develop a full inference pipeline across various edge
devices, choosing the best inference strategy depending on hardware and network
conditions, minimizing inference latency through the use of custom network
protocols, and improving ease of use for software developers that wish to use
collaborative intelligence in their own applications.

This chapter looks at how to bring collaborative intelligence out of the
research lab and into the real world. First, we will discuss a proof-of-concept
implementation that was developed for Android, and an inference pipeline that
allows interchangeability between its various components. Next, we will
describe how total inference latency varies depending on the model, strategy,
hardware performance, and network conditions. Then, we will propose a blueprint
for a network protocol customized for transmitting real-time tensor streams,
keeping in mind latency, throughput, and backpressure. Lastly, we will look at
some library code that was developed to help conduct research and also
functioned as part of the proof-of-concept.

\section{Prototype}
\label{sec:implementation/prototype}

\newcommand{\makeurl}[1]{%
  \url{https://github.com/YodaEmbedding/collaborative-intelligence#1}%
}
\newcommand{\urlandroid}{\makeurl{/tree/thesis/android}}
\newcommand{\urlserver}{\makeurl{/blob/thesis/server.py}}
\newcommand{\urlmonitor}{\makeurl{/tree/thesis/server-monitor}}

\begin{figure}
  \centering
  \includegraphics[width=0.7\textwidth]{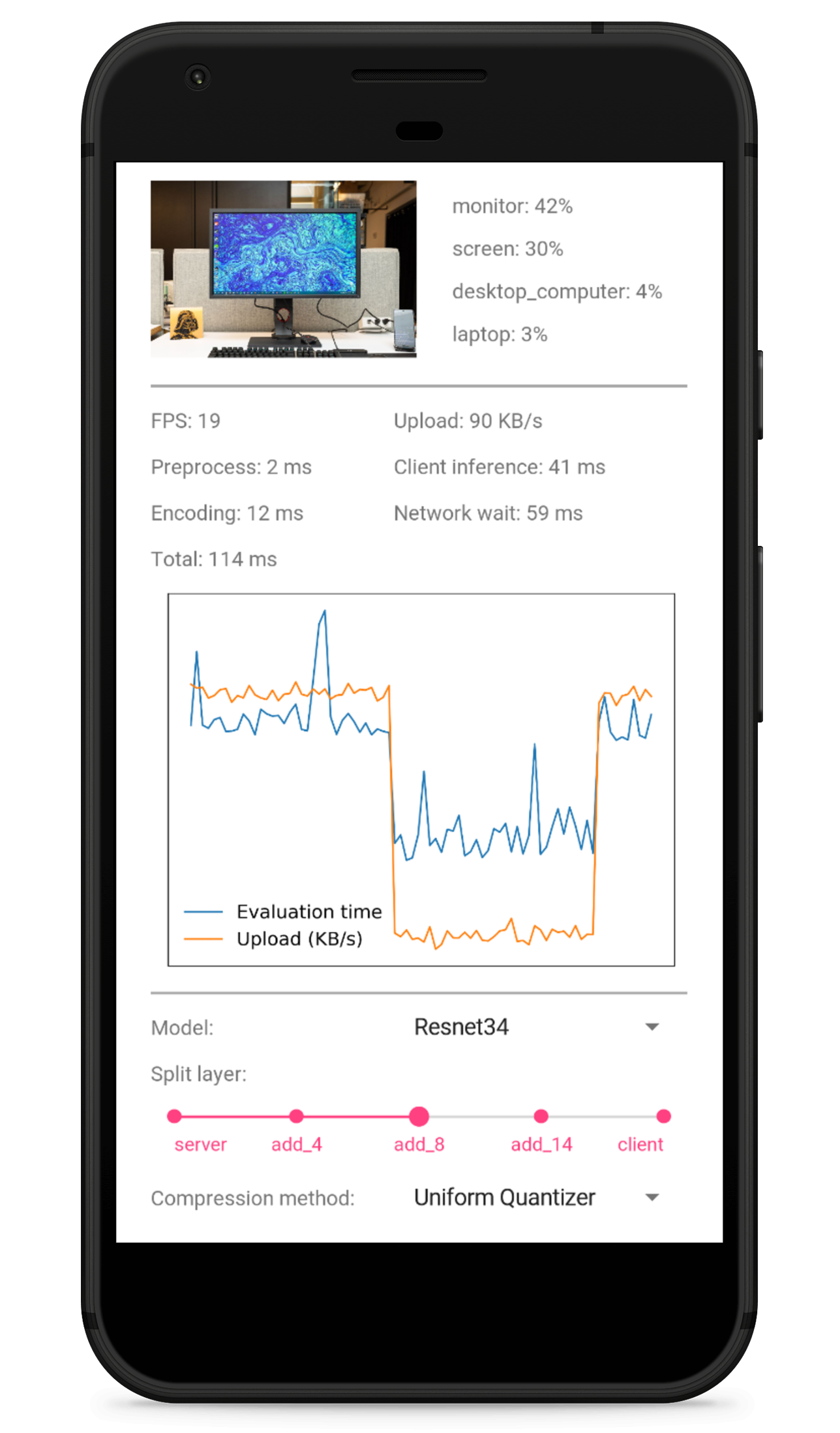}
  \caption[Android app]{%
    Demo app running on Android device. One may choose between the model, layer
    split points, feature tensor compression method, and also control the
    available upload bandwidth. Plots total inference time and
    serialized/compressed tensor upload size for each frame.%
  }
  \label{fig:android_app}
\end{figure}

A complete prototype was designed to demonstrate and help compare between
server-only, client-only, and shared inference strategies for a variety of
models. To function as the mobile edge client, an Android
app\footnote{\urlandroid{}} was written in Kotlin and RenderScript, utilizing
the TensorFlow Lite, Fotoapparat, and RxJava libraries. A screenshot of the app
in action is shown in \cref{fig:android_app}. To perform the server-side
inference, a server\footnote{\urlserver{}} was written in Python to receive and
process inference requests. Finally, a GUI app for desktop or web
clients\footnote{\urlmonitor{}} was written in TypeScript, NodeJS, and the
Electron framework to display a preview of the server's reconstructed tensor,
debugging information, and any other additional real-time statistics. The
models available include various ResNet models (18, 34, 50, 101, 152) and VGG,
though one can add additional models by splitting the model, converting it to
\texttt{.tflite} format, and adding the relevant entries to
\texttt{models.json} (see repository for details).

\subsection{NeurIPS demo}

\begin{figure}
  \centering
  \includegraphics[width=.8\linewidth]{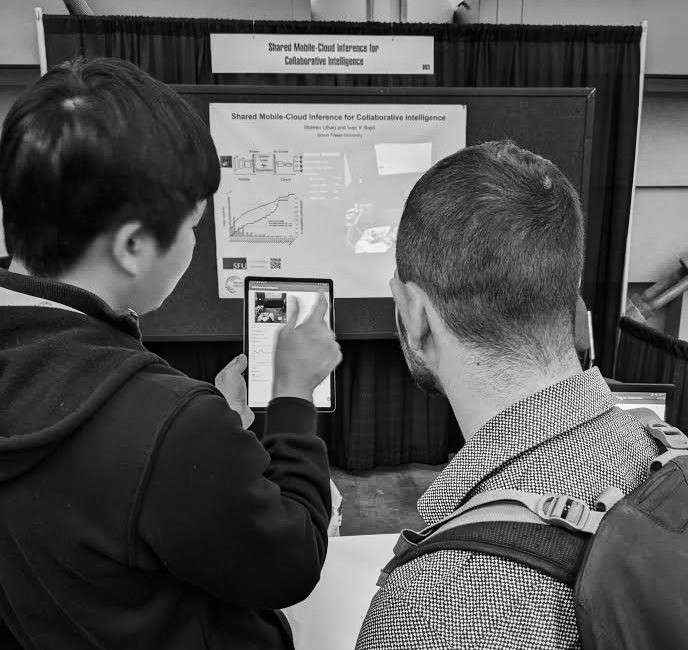}
  \caption[Demo at NeurIPS 2019]{%
    Live demo of prototype at NeurIPS 2019.%
  }
  \label{fig:demo}
\end{figure}

The prototype was demonstrated live at the NeurIPS 2019 conference. A paper
describing the demo is available in~\cite{ulhaq2020shared}. Two Android devices
were used as clients; one client was connected to a nearby server on a LAN, and
the other client was connected to a remote server in a nearby city. Each server
utilized an NVIDIA GeForce GTX Titan X GPU, whereas the clients utilized the
Qualcomm Snapdragon 845 and Qualcomm Snapdragon 855 SoCs. \cref{fig:demo}
contains a photo of the interactive demo booth from this event.

\subsection{Inference pipeline}

\begin{figure}
  \centering
  \tikzstyle{block} = [
  rectangle,
  fill=cyan!20!white,
  minimum height=4em,
  rounded corners,
  text centered,
  text width=6.5em
]

\tikzstyle{terminal} = [
  ellipse,
  fill=red!60!blue!20!white,
  minimum height=4em,
  text centered,
  text width=4.5em
]

\tikzstyle{arrow} = [draw, -latex, very thick, color=black!35]
\tikzstyle{bracestyle} = [draw, very thick, color=black!35]
\tikzstyle{divider} = [draw, dotted, very thick, color=black!35]
\tikzstyle{label} = [color=black!70]
\tikzstyle{bracelabel} = [label, font=\large, rotate=90, anchor=south]
\tikzstyle{smalllabel} = [color=black!60, font=\small]

\begin{tikzpicture}[node distance=2.5cm, auto]
  \node at (-3, 1) {};
  \node at (3, -16) {};

  \node[terminal] (input) {Input signal};
  \node[block, below of=input] (preprocess) {Preprocess};
  \node[block, below of=preprocess] (inference_client) {Client inference};
  \node[block, below of=inference_client] (postencoder) {Postencoder};
  \node[block, below of=postencoder] (predecoder) {Predecoder};
  \node[block, below of=predecoder] (inference_server) {Server inference};
  \node[terminal, below of=inference_server] (output) {Output signal};

  \path[arrow] (input) -- (preprocess);
  \path[arrow] (preprocess) -- (inference_client);
  \path[arrow] (inference_client) -- (postencoder);
  \path[arrow] (postencoder) -- (predecoder);
  \path[arrow] (predecoder) -- (inference_server);
  \path[arrow] (inference_server) -- (output);

  \node (network_a_west) at ($(postencoder.west)!0.5!(predecoder.west)$) {};
  \node (network_a_east) at ($(postencoder.east)!0.5!(predecoder.east)$) {};
  \node[smalllabel] (network_a) at ($(network_a_west) - (1, 0)$) {Network};
  \path[divider] ($(network_a_west)$) -- ($(network_a_east)$);

  \node (network_b_west) at ($(inference_server.west)!0.5!(output.west)$) {};
  \node (network_b_east) at ($(inference_server.east)!0.5!(output.east)$) {};
  \node[smalllabel] (network_b) at ($(network_b_west) - (1, 0)$) {Network};
  \path[divider] ($(network_b_west)$) -- ($(network_b_east)$);

  \node (client_a) at ($(-0.75, 0) + (preprocess.west)$) {};
  \node (client_b) at ($(-0.75, 0) + (postencoder.west)$) {};
  \node[bracelabel] at
    ($(client_a)!0.5!(client_b) + (-0.40, 0)$) {Client};
  \path[bracestyle]
    ($(client_a) + (0.25, 0)$) --
    ($(client_a)$) --
    ($(client_a)!0.5!(client_b)$) --
    ($(client_a)!0.5!(client_b) + (-0.25, 0)$) --
    ($(client_a)!0.5!(client_b)$) --
    ($(client_b)$) --
    ($(client_b) + (0.25, 0)$);

  \node (server_a) at ($(-0.75, 0) + (predecoder.west)$) {};
  \node (server_b) at ($(-0.75, 0) + (inference_server.west)$) {};
  \node[bracelabel] at
    ($(server_a)!0.5!(server_b) + (-0.40, 0)$) {Server};
  \path[bracestyle]
    ($(server_a) + (0.25, 0)$) --
    ($(server_a)$) --
    ($(server_a)!0.5!(server_b)$) --
    ($(server_a)!0.5!(server_b) + (-0.25, 0)$) --
    ($(server_a)!0.5!(server_b)$) --
    ($(server_b)$) --
    ($(server_b) + (0.25, 0)$);
\end{tikzpicture}
  \caption[Full inference pipeline]{%
    Full inference pipeline. An input signal is preprocessed (e.g. an image is
    cropped, resized, and normalized). Then, it goes through the first half of
    the inference model on the client. The intermediate tensor is also passed
    through a simple GPU-based encoder (e.g. an 8-bit quantizer) attached to
    the end of the client-side model. This tensor is then more heavily
    compressed by the \emph{postencoder}, which may perform its operations on
    the CPU or on a dedicated hardware chip for codecs. This outputs a
    serialized stream of bytes that can be transmitted over the network. The
    server receives the byte stream and feeds it into the \emph{predecoder},
    which deserializes the byte stream into a tensor. This tensor is then
    inputted into the server inference model, which is prefixed by a decoder
    that reconstructs an approximation of the original tensor. Finally, the
    resulting output signal is serialized into a byte stream and transmitted
    back to the client.%
  }
  \label{fig:pipeline}
\end{figure}
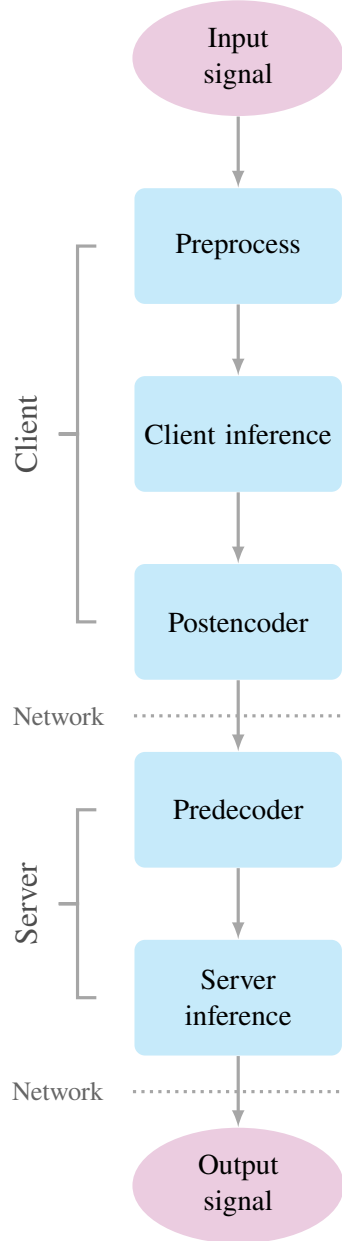

\cref{fig:pipeline} shows a flowchart overview of the complete inference
pipeline. The pipeline takes an input signal, preprocesses it, runs it through
the client-side inference model, applies encoding on partially inferred tensor
(on the GPU), serializes the resulting tensor via a \emph{postencoder} (which
runs on the CPU or other dedicated hardware), transmits the data over the
network, deserializes with a corresponding \emph{predecoder}, applies decoding
on the deserialized tensor (on the GPU), runs the reconstructed tensor through
the server-side inference model, and finally sends the result of the complete
inference back to the client.

The client begins by notifying the server which model to use. When the client
needs to initialize or change the model and other configuration settings, a
\texttt{ModelSwitchRequest} is made. This sends over details including the
model, split layer, encoder, and postencoder. Once the inference pipeline is
configured and ready, the server replies with a \texttt{ModelReadyResponse}.
This ensures that the model and postencoder are preloaded and that the
perceived server inference latency is minimized for when the client sends its
first inference request to the server.

Frames that are inputted into the pipeline often require preprocessing.
RenderScript is used to preprocess the input frames efficiently. The camera
input is converted from YUV to RGB, cropped to 1:1, resized to $224 \times
224$, rotated to an upright orientation, and converted to float, before being
fed into the client-side model. The reshaping/tiling code could also be
converted to RenderScript for better performance, though the current
Kotlin-only implementation is quick enough for small tensor sizes.

The encoder is attached to the end of the client-side model, and is a part of
the deep model itself. The encoder can be a trainable network, such as the
reduction unit of a "butterfly unit"~\cite{eshratifar2019collaborative}, which
reduces the size of the tensor. Or it can simply be a quantizer, handling
data-heavy computations directly with a GPU, before the data is transferred to
RAM. For instance, the \texttt{UniformQuantizationU8Encoder} layer uniformly
quantizes the floating-point values of the preceding layer to an 8-bit
\texttt{uchar} over a fixed clipping range.

After the encoder runs, the tensor output is copied from GPU memory to RAM.
From here, it is fed into the postencoder, which runs on the CPU or a dedicated
hardware chip (e.g. H.264 ASIC). The postencoder is a codec such as JPEG, PNG,
H.264, or some other custom codec. The postencoder handles any necessary
reshaping and tiling of the tensor, and outputs a compressed byte stream that
can be transmitted over the network; in the current implementation, this is
done over TCP. Analogous to the client-side postencoder, there is a server-side
predecoder, which decodes the byte stream, and feeds the resulting tensor into
the server-side model. Attached to the beginning of the server-side model is a
decoder corresponding to the encoder at the end of the client-side model. This
decoder reconstructs the tensor to be of the correct data type and shape,
before propagating the tensor through the rest of the model.

The prediction tensor outputted from the server-side inference can be sent back
to the client, optionally after a further processing/decoding step. In the case
of the demo app and server, the probability vector is decoded into the top-n
labels and probabilities. Along with the request id and measured server-side
inference latency (useful for rate limiting and monitoring statistics), these
are sent back in a \texttt{ResultResponse}. An example JSON serialization of
this is given:

\vspace{\baselineskip}
\noindent
\begin{minipage}{\linewidth}
  \begin{codelisting}{json}
{
  "frameNumber": "<int>",
  "inferenceTime": "<int>",
  "predictions": {
    "label": {"name": "<str>", "description": "<str>", "score": "<int>"},
    "label": {"name": "<str>", "description": "<str>", "score": "<int>"},
    ...
  }
}
  \end{codelisting}
\end{minipage}
\vspace{0.5\baselineskip}

To prevent overloading the network with too many write requests, no new
inference requests are made by the client until a \texttt{ConfirmationResponse}
(containing a unique frame request id and the number of bytes received) is
received from the server.

\section{Latency model}
\label{sec:implementation/latency_model}

The total inference latency for a shared inference strategy can be expressed as
a sum of various parameters:
\begin{equation}
  I_t =
  \underbrace{
    \vphantom{\frac{D}{B}}
    I_c + E_c
  }_{\text{client-side}}
  +
  \underbrace{
    \frac{D}{B} + \textrm{RTT}
  }_{\text{network}}
  +
  \underbrace{
    \vphantom{\frac{D}{B}}
    E_s + I_s
  }_{\text{server-side}}
  + \epsilon
  \label{eq:latency/total}
\end{equation}
where
\begin{conditions}
  I_t & total inference time \\
  I_c & inference time of client-side model \\
  I_s & inference time of server-side model \\
  E_c & serialization/encoding time for client output tensor \\
  E_s & deserialization/decoding time for server input tensor \\
  D & size of serialized/compressed tensor data \\
  B & rate of data transfer (bandwidth) \\
  \textrm{RTT} & round trip time \\
  \epsilon & other latencies (negligible) \\
\end{conditions}

\newcommand{\pidx}[1]{{\, (#1)}}

The values of these parameters vary wildly depending on the hardware, network
conditions, the model, choice of split layer, and available compression
options. A particular strategy $\Omega^\pidx{l}$ can be associated with a
4-tuple $\left( I_c^\pidx{l}, \, E_c^\pidx{l}, \, E_s^\pidx{l}, \, I_s^\pidx{l}
\right)$. Then we may write the total inference latencies for the various
strategies as:
\begin{alignat}{8}
  I_t^\pidx{l}
    &= I_c^\pidx{l} & {} + {}
    & E_c^\pidx{l}
    && + E_s^\pidx{l}
    && + I_s^\pidx{l}
    && + \textrm{RTT}
    && + \frac{D^{\pidx{l}}}{B}
    && = b^\pidx{l} + \frac{D^{\pidx{l}}}{B}
    & \quad \quad \text{split layer $l$}%
    \\%
  I_t^\pidx{s}
    & = &
    & E_c^\pidx{s}
    && + E_s^\pidx{s}
    && + I_s^\pidx{s}
    && + \textrm{RTT}
    && + \frac{D^{\pidx{s}}}{B}
    && = b^\pidx{s} + \frac{D^{\pidx{s}}}{B}
    & \quad \quad \text{server-only}%
    \\%
  I_t^\pidx{c}
    & = I_c^\pidx{c} &
    &
    &&
    &&
    &&
    &&
    && = b^\pidx{c}
    & \quad \quad \text{client-only}%
    \vphantom{\frac{D^(c)}{B}}%
  \label{eq:latency/reciprocal_bandwidth}
\end{alignat}

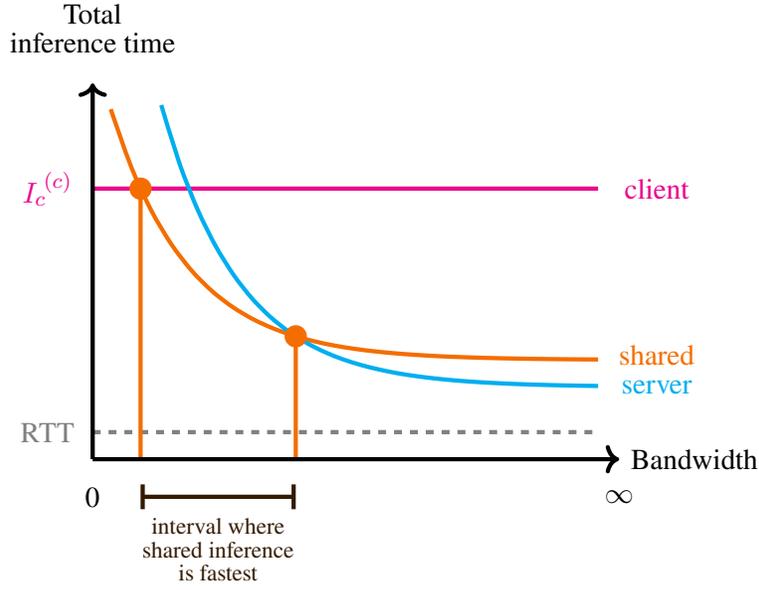
\begin{figure}
  \centering
  \def\sc{1.2}

\tikzstyle{client} = [magenta]
\tikzstyle{shared} = [orange!95!black!90!red]
\tikzstyle{server} = [cyan]
\tikzstyle{rtt} = [gray, dashed]
\tikzstyle{intervalcol} = [orange!20!black]

\tikzstyle{arrow} = [
  ->,
  ultra thick,
]

\tikzstyle{label} = [
  inner ysep=1em,
  align=center,
]

\tikzstyle{funcline} = [
  ultra thick,
  smooth,
]

\tikzstyle{function} = [
  funcline,
  scale=\sc,
  variable=\x,
]

\tikzstyle{dot} = [
  circle,
  fill,
]

\tikzstyle{interval} = [
  intervalcol,
  scale=0.8,
  align=center,
]

\def\rtt{0.3}
\def\dc{0.0} \def\bc{3.0}
\def\dl{3.5} \def\bl{1.1}
\def\ds{7.5} \def\bs{0.8}

\FPeval{\xc}{(2 * ln((\dc - \dl) / (\bl - \bc)) / ln(10))}
\FPeval{\xs}{(2 * ln((\ds - \dl) / (\bl - \bs)) / ln(10))}
\FPeval{\yc}{(\dl * pow(-\xc / 2, 10) + \bl)}
\FPeval{\ys}{(\dl * pow(-\xs / 2, 10) + \bl)}


\begin{tikzpicture}
  \renewcommand{\baselinestretch}{0.8}

  \draw[function, client, domain=0.00:2.8] plot (2 * \x, {\dc * 10^-\x + \bc});
  \draw[function, shared, domain=0.10:2.8] plot (2 * \x, {\dl * 10^-\x + \bl});
  \draw[function, server, domain=0.38:2.8] plot (2 * \x, {\ds * 10^-\x + \bs});
  \draw[function, rtt,    domain=0.00:2.8] plot (2 * \x, \rtt);

  \node[client] at (7.5, 3.60) {client};
  \node[shared] at (7.5, 1.38) {shared};
  \node[server] at (7.5, 0.96) {server};

  \filldraw[shared] ({\xc * \sc}, {\yc * \sc}) circle (4pt);
  \filldraw[shared] ({\xs * \sc}, {\ys * \sc}) circle (4pt);

  \draw[funcline, shared] ({\xc * \sc}, {\yc * \sc}) -- ({\xc * \sc}, 0);
  \draw[funcline, shared] ({\xs * \sc}, {\ys * \sc}) -- ({\xs * \sc}, 0);
  \draw[funcline, intervalcol, |-|] ({\xc * \sc}, -0.5) -- ({\xs * \sc}, -0.5);
  \node[interval] at ({(\xc + \xs) / 2 * \sc}, -1.2)
    {interval where \\ shared inference \\ is fastest};

  \node[rtt]    at (-0.6, {\rtt * 1.2}) {RTT};
  \node[client] at (-0.6, {\bc  * 1.2}) {$I_c^{\; (c)}$};
  \node at (0.0, -0.5) {0};
  \node at (7.0, -0.5) {$\infty$};

  \draw[arrow] (0, 0) -- (0, 5) node[label, above] {Total \\ inference time};
  \draw[arrow] (0, 0) -- (7, 0) node[label, right] {Bandwidth};
\end{tikzpicture}
  \caption[Total inference latency vs bandwidth]{%
    Total inference latency versus bandwidth for the given conditions
    \cref{eq:latency/conditions}. Lower is better. At zero bandwidth,
    client-only inference performs best. However, shared inference becomes
    dominant as bandwidth increases, until finally, server-only inference
    surpasses shared inference due to faster server-side hardware and the
    benefits of transmitting less data becoming less significant.
  }
  \label{fig:latency_vs_bandwidth}
\end{figure}

To compare a particular shared inference strategy with a client-only or
server-only inference strategy, or even other shared inference strategies, we
can fix the round trip time ($\textrm{RTT}$) across strategies and vary
bandwidth ($B$). Then, if the following conditions
\begin{gather}
  0 = D^\pidx{c} < D^\pidx{l} < D^\pidx{s} \\
  b^\pidx{s} < b^\pidx{l} < b^\pidx{c}
  \label{eq:latency/conditions}
\end{gather}
are satisfied, a plot of the total inference time ($I_t$) versus bandwidth may
appear similar to \cref{fig:latency_vs_bandwidth}. For some interval near zero
bandwidth, the client-only strategy is to be preferred. However, as more
bandwidth becomes available, the shared inference strategy begins to dominate.
Then, as bandwidth further increases, the server-only strategy begins to
dominate. Thus, there are regions of operation in which a particular strategy
is preferable.

\begin{figure}
  \centering
  \includegraphics[width=\linewidth]{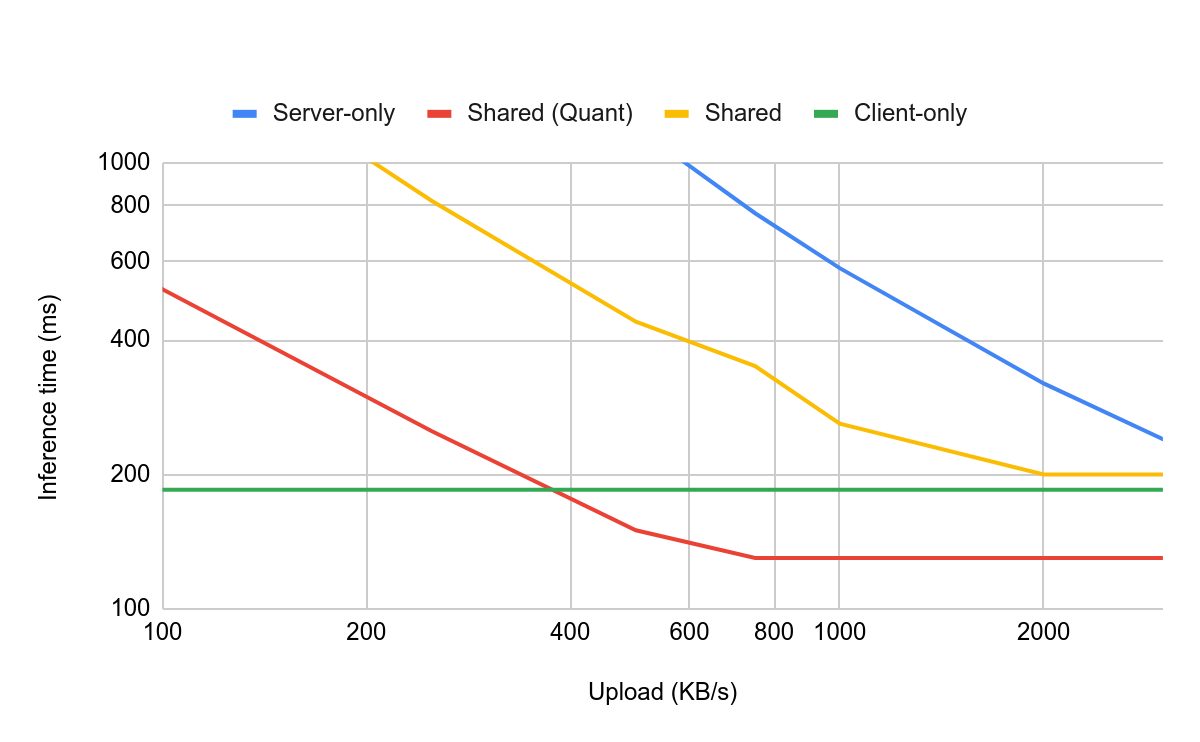}
  \caption[Total inference latency vs bandwidth (actual)]{%
    Total inference latency versus bandwidth. This data was collected using the
    prototype described in \cref{sec:implementation/prototype}. The model under
    test is ResNet-34. The data transmitted in this experiment is uncompressed.
    The mobile client is a Samsung Galaxy S10 phone (Qualcomm Snapdragon 845)
    that is connected to the internet via a Wi-Fi network with a maximum
    available upload bandwidth of 3 MB/s. The server utilizes an NVIDIA GeForce
    GTX Titan X GPU. There is a distance of less than 5 km between the client
    and server, with an average ping time of 5 ms. To simulate different
    available bandwidths, the outgoing traffic is artificially rate limited
    within the client software itself.%
  }
  \label{fig:measurements}
\end{figure}

This model of the latency is supported by experiments that were conducted on
the prototype described in \cref{sec:implementation/prototype}.
\cref{fig:measurements} shows how total inference time varies as the client's
upload bandwidth changes. Though the server-only inference has not reached a
horizontal asymptote at the maximum bandwidth capacity, it is expected that it
should continue to decrease until it reaches some value underneath the
quantized shared inference curve.

As network conditions such as round trip time and bandwidth availability
fluctuate, one can switch between different strategies to maintain a low total
inference latency. In practice, one should apply hysteresis to the decision of
switching between strategies. This is because switching models consumes time
and energy as the models are loaded into memory. Furthermore, if network
conditions are highly unstable, it is best to stick to a single model that
reduces the average latency and also has a large enough region of operation for
which the total inference latency is tolerable.

If one also considers the strategies for different split layers, it makes sense
to pick a small subset of split layers that operate tolerably for a variety of
network conditions. Since TensorFlow Lite client-side models must be stored
individually in \texttt{.tflite} files, and --- as of the time of this writing
--- since subgraphs cannot be extracted from a single trained model, it is
important to restrict the number of client-side models stored so that they
don't consume unnecessary storage space on the mobile device.

\section{Network protocol}
\label{sec:implementation/network_protocol}

To avoid congestion and improve flow control, TCP often makes use of Nagle's
algorithm~\cite{rfc896}. Nagle's algorithm forces the TCP socket to wait for an
ACK response before transmitting more data. This introduces latency due to
round-trip times. Thus, Nagle's algorithm should be disabled for large writes;
instead, it is better to buffer the data when writing. One can set
\texttt{TCP\_NODELAY} on a Linux socket to disable Nagle's algorithm.

Instead of using TCP for transmission of tensor data, one could theoretically
achieve lower latencies by using a custom protocol built on top of UDP. This
would also allow the usage of custom error correction methods designed with
compressed tensor data in mind.

A better real-time inference protocol design for minimizing inference latency
and maximizing throughput should send maximally sized packets of tensor data,
unless the end of a given tensor's serialized byte stream is reached. In that
case, it should immediately flush the remaining contents of the send buffer and
not wait for the send buffer to be filled. In pseudocode:

\begin{codelisting}{python}
def process_send_buffer():
  if send_buffer.is_empty():
    return

  if send_buffer.size() >= mss:
    data = send_buffer.pop_bytes(mss)
    send(Packet(data))
    return

  if send_buffer.contains_end_of_tensor():
    data = send_buffer.pop_until_end_of_tensor()
    send(Packet(data))
\end{codelisting}

\noindent
where MSS (Maximum Segment Size) is the maximum amount of data that can be sent
in a packet, excluding headers.

Another issue arises when the client attempts to send more requests than the
server can process or that bandwidth of the connection can support. This
phenomenon is often dubbed \emph{backpressure}. To avoid backpressure, one
needs to estimate expected server inference times as well as the available
bandwidth of the connection. These estimates are to be utilized at the
application layer of the Open System Interconnection (OSI) model to limit the
client from sending requests when it is unwise to do so without increasing
latencies or reducing overall throughput.

Server inference times can be tricky to determine if the server is handling
multiple requests from different clients at once, particularly if the server
implements batching to maximize throughput, or if the inference times are
non-equal. A typical strategy is that the server manages its own time slicing
and notifies its clients about their particular rate limits, which each client
adheres to (and if a malicious client \emph{doesn't} adhere to its given rate
limits, hard limits can still be enforced server-side).

Because the available bandwidth is non-deterministic and varies over time,
bandwidth limits can be tricky to estimate. One proposed method for developing
a bandwidth estimation scheme involves the server sending confirmation upon the
reception of a chunk of tensor data. Note that this confirmation does not
necessarily correspond to "ACK". If an out-of-order protocol such as UDP is
used, the server should reply with a confirmation message containing
information on the total number of bytes received, a unique identifier for the
packet, and the time it is received at (assuming synchronized clocks). This
helps the client's bandwidth model account for out-of-order packets and
potentially lost packets (for which it should also maintain some probabilistic
model). By keeping a history of the confirmations, and also accounting for the
additional latency introduced by the ping or round trip time (RTT), one can
develop a real-time estimate of the bandwidth.


One simple strategy to avoid backpressure in the shared inference pipeline from
the application layer of the OSI model is given as follows:

\begin{codelisting}{python}
def send_request_to_server(request):
  while True:
    # Wait until server rate limit is satisfied.
    if now() < last_request_time + server_rate_limit:
      Thread.sleep(0)
      continue

    # Wait until we expect not to exceed bandwidth limits.
    if estimate_unreceived_bytes() > 0:
      Thread.sleep(0)
      continue

    write(request.data)
    break
\end{codelisting}

For real-time processing of input frames, the aforementioned strategy is not
sufficient: one needs to also account for client-side inference times. To
minimize total inference latency, one strategy is to drop frames if the rate
limiter can predict that any tensor data that is currently being sent to the
network will not be flushed by the time the current frame completes
preprocessing, client-side inference, and compression. This also reduces the
wastage of client-side computational resources; frames no longer need to be
dropped \emph{after} spending time on client-side computation, only to realize
that there is no remaining bandwidth or server availability. An algorithm for
real-time frame processing, optimized for latency, is given as follows:

\begin{codelisting}{python}
def should_process_frame(frame):
  client_remain = estimate_client_process_latency()
  server_remain = last_request_time + server_rate_limit - now()
  bandwidth_remain = estimate_unreceived_bytes() / estimate_bandwidth()

  # Drop frame if server or bandwidth won't be available in time
  if (client_remain < server_remain or client_remain < bandwidth_remain):
    return False

  return True
\end{codelisting}

Throughput (the total number of frames processed) can be further increased
through a more advanced algorithm; however, any such algorithm will involve
latency tradeoffs.

\section{Library}

To help aid in analysis and determining good split points, commonly needed
functionality (splitting, inference, tiling, encoding, statistics) was
extracted into a custom python module%
\footnote{%
  \url{
    https://github.com/YodaEmbedding/collaborative-intelligence/tree/master/src/lib
  }%
}
for collaborative intelligence. This section contains some usage examples.

Begin by importing relevant modules:

\begin{codelisting}{python}
import collaborativeintelligence as ci
import numpy as np
from tensorflow import keras
\end{codelisting}

Next, load the required \texttt{model}, and create the \texttt{encoder} and
\texttt{decoder} layers, which will be attached to the end of the client model
and beginning of the server model, respectively. For example, we can load a
\texttt{keras.Model} from \texttt{\textquotedbl filename.h5\textquotedbl} and
attach a simple uniform 8-bit quantization layer as the encoder layer:

\begin{codelisting}{python}
model = keras.models.load_model("filename.h5")
encoder = ci.UniformQuantizationU8Encoder(clip_range=[-3., 3.])
decoder = ci.UniformQuantizationU8Decoder(clip_range=[-3., 3.])
\end{codelisting}

To split the model:

\begin{codelisting}{python}
model_client, model_server, _ = ci.split_model(
    model, layer=layer_name, encoder=encoder, decoder=decoder
)
\end{codelisting}

Now, construct the \texttt{postencoder} and \texttt{predecoder}:

\begin{codelisting}{python}
shape = model_client.output_shape[1:]
dtype = model_client.dtype
tensor_layout = ci.TensorLayout.from_shape(shape, "hwc", dtype)
postencoder = ci.JpegPostencoder(tensor_layout, quality=20)
tiled_layout = postencoder.tiled_layout
predecoder = ci.JpegPredecoder(tiled_layout, tensor_layout)
\end{codelisting}

Finally, we may use the entire pipeline on a series of \texttt{frames} as
follows%
\footnote{%
  The single tensor \texttt{x} is put into a one-item batch preceding
  prediction via \texttt{x[np.newaxis]}. Then, we extract the first and only
  item from the predicted batch via \texttt{predict(\ldots)[0]}. Here, we avoid
  collecting and batching more tensors because we are prioritizing low latency
  in lieu of throughput.%
}%
:

\begin{codelisting}{python}
for frame in frames:
    x = frame
    x = model_client.predict(x[np.newaxis])[0]
    x = postencoder.run(x)
    x = predecoder.run(x)
    x = model_server.predict(x[np.newaxis])[0]
    print(x)
\end{codelisting}

\section{Summary}

In this chapter, we discussed the technologies and design choices used in
developing a proof-of-concept Android app, which was demoed at the NeurIPS 2019
conference. We established a concrete inference pipeline that took a real-time
inference signal, preprocessed it, ran client-side inference, further
compression-related tasks in "postencoder" separately from the deep model
itself, sent the resulting data over the network, reassembled the data and fed
it through a server-side "predecoder", and finally ran the remaining deep model
inference on the server.

Next, we established a model for assessing latency of various inference
strategies using estimated hardware latencies and network conditions. We
showed, in terms of inference latency, that certain strategies are optimal
within certain intervals of available bandwidth. Of course, one may
alternatively seek to optimize for a combination of other variables including
throughput, energy usage, bandwidth costs, or data privacy --- these will
produce different characteristic curves.

Next, we proposed a blueprint for a network protocol customized for
transmitting real-time tensor streams. Though TCP is a reliable protocol for
guaranteeing transmission of data, we can do better by building upon UDP in
order to improve latency -- particularly since exact and complete reception of
tensor data is not necessary to perform a reasonably accurate inference, as was
discussed in \cref{sec:tensor_streams/error_concealment}.

Finally, we briefly looked at some functionality provided by a library that was
developed to help conduct experiments for this thesis, and also was used as
part of the proof-of-concept server implementation.

\chapter{Conclusion}

\section{Thesis summary}

Collaborative intelligence is a new concept for improving the mobile experience
and bringing more powerful AI models to edge devices. This thesis looked to
provide further insight into developing the shared inference strategy and
provide pathways into developing practical applications that utilize this
technique.

One of the core prerequisites towards making collaborative intelligence
effective involves tensor data compression. In
\cref{sec:introduction/compressibility}, we showed that information theory
supports the idea that partial processing of a data tensor via non-generative
feedforward networks reduces its entropy, and thus its theoretically minimum
losslessly compressed size.

In \cref{chap:compression}, we looked at methods for improving this key
requirement. By quantizing BatchNorm layers or layers that follow batch
normalized layers to as few as 6 quantization bins, one can often achieve
near-equal Top-1 accuracies in image classification networks such as ResNet.
This gave us a compression ratio of 12:1. To improve on this, we next looked at
leveraging existing image codecs such as JPEG and JPEG 2000. These generated
rate-accuracy curves that improved in shared inference accuracy for low rates
as we split deeper into the model. For instance, at the \texttt{add\_7} layer
of ResNet-34, a compressed feature tensor size of 9 KB via JPEG resulted the
same accuracy drop of 1\% as a 15 KB JPEG compression of the $224 \times 224
\times 3$ input image.

In \cref{chap:tensor_streams}, we investigated the consequences of modifying
and transforming values of the input and intermediate tensors. We discovered
that we can minimize the drop in accuracy due to missing tensor data by filling
in missing tensor entries with their precomputed mean value over a sample set
of images. We also saw that as we apply common video-like transformations to
the input data, such as the translation of an image, the feature tensor of a
CNN responds in the same way. In fact, by precomputing the global motion vector
between the reference input image and a translated input image, we can
accurately predict the entries of their corresponding feature tensors to a high
degree of accuracy by applying a downscaled version of the motion vector, where
the tensor entries are interpolated if necessary. These ideas can be used to
develop compression schemes for streams of tensors that result from a sequence
of input data that contains temporal redundancies (e.g. video).

In \cref{chap:implementation}, we took some of the concepts discussed in prior
chapters and used them to develop an Android app and server for a working
demonstration of collaborative intelligence. We also discussed possible
extensions and improvements to this implementation, including a network
protocol for real-time tensor streams. Finally, we looked at the range of
available network bandwidths for which a shared inference strategy can perform
better than a server-only or client-only inference strategy.

\section{Future work}

\subsection{Model architecture}

To expand the scope of collaborative intelligence, it is important to
incorporate collaborative intelligence techniques with a larger number of
existing models. One can also adapt and modify existing models to work better
with these techniques by inserting new layers. For instance, the insertion of
BatchNorm layers can be used to constrain the range of values and reduce the
effect on inference accuracy due to lossy reconstructions. Encoder/decoder
layers such as butterfly units~\cite{eshratifar2019collaborative} can also be
useful in reducing tensor dimensionality. Training with compressible features has also been proposed~\cite{saeed_icip_2019}. Finally, more work needs to be done
to design and introduce new model architectures that are specifically designed
for collaborative intelligence, and exhibit the characteristics described in
\cref{sec:introduction/choosing_cut}.

\subsection{Tensor compression}

Compression of feature tensors extracted from deep models is an active area of
research~\cite{choi_icip_2018,choi_mmsp_2018,choi_icassp_2020,bob_icme_2020,chen_etal_TIP_2020}.
For CNN architectures, the primary concern is exploiting spatial (intraframe)
and temporal (interframe) redundancies within a tensor stream that is generated
from an input video stream. For this, it is useful to utilize existing video
codecs for practical reasons such as reduced development effort, existing
optimized implementations, and availability of hardware optimizations for
operations used in encoding and decoding. Open source implementations (e.g. the
x264 encoder for H.264) can be modified to assist the encoding process.
Furthermore, one can also modify both the encoder \emph{and} decoder to
compress to a custom format which extends or modifies the original
specification.

One possible modification makes use of the idea that we may reuse the same set
of motion vectors across all the channels (as demonstrated in
\cref{sec:tensor_streams/input_transformations}). However, the H.264 format
requires us to store a copy of this set of motion vectors for each channel. To
prevent this duplication from occurring, we must modify the encoder to store
only one set of motion vectors in a custom H.264-like format, and the decoder
to copy these motion vectors across all channels.

Furthermore, the typical coding block sizes are powers of 2, whereas channel
dimensions may not be%
\footnote{%
  In CNNs, channels are frequently sized at integer multiples of $7 \times 7$.%
}%
. Non-padded tiling of these channels makes it difficult to exploit the
periodicity among the channels that was noted in \cref{eq:cnn_redundancies}.
DCT in particular produces different coefficients for phase shifted data. This
deficiency needs to be addressed in order to achieve significantly lower
bitrates.

\subsection{Error concealment}

There is often much spatially and temporally redundant information carried
within tensor streams. Techniques like the ones mentioned in
\cref{sec:tensor_streams/error_concealment} use prior statistical knowledge,
whereas tensor completion methods exploit more sophisticated spatial
redundancies within tensors. However, there are also temporal redundancies that
have not yet been considered. For instance, in real-time video transmission,
one can fill in lost motion vectors using the motion vectors of previous
frames. Similar or alternative techniques could also be developed for tensor
streams.

\subsection{Network protocol}

As described in \cref{sec:implementation/network_protocol}, a custom protocol
for real-time transmission of compressed tensor streams would allow for lower
latencies than the current reliable but slower TCP implementation. To deal with
lost or corrupted packets, such a protocol could also implement error
concealment methods designed specifically for tensor data.

\subsection{Libraries}

To make collaborative intelligence techniques easier to implement and design,
and increase accessibility for non-experts, it is useful to have a set of
libraries. These libraries can be utilized by both collaborative intelligence
researchers as well as software developers of mobile apps and server back-ends,
who may desire practical production grade solutions. Such libraries could
contain tools for analyzing, developing, tuning, and deploying collaborative
intelligence strategies. Useful features to consider include:

\begin{itemize}
  \item backpressure-aware shared inference pipelines for real-time data
    processing
  \item combinators for methods used in tensor compression (e.g. quantization,
    tiling/weaving to convert between 3D and 2D tensor shapes, and interfacing
    with image/video/tensor codecs)
  \item implementation of a real-time tensor streaming protocol
  \item monitoring and collection of statistics for analysis
  \item real-time automated transitioning to optimal inference strategy based
    on network conditions
  \item simulations for assessing performance under different network conditions
  \item utilities for determination of good split points
\end{itemize}

\backmatter

\clearpage
\phantomsection
\addcontentsline{toc}{chapter}{References}
\printbibliography[title=References]

\appendix
\chapter{Additional figures}
\label{sec:appendix/additional_figures}

\providecommand{\fig}[5]{}
\renewcommand{\fig}[5]{%
  \begin{subfigure}[t]{.5\textwidth}
    \centering
    \includegraphics[width=\linewidth]
    {img/accuracyvskb/jpeg_uniquant256/\modelname/\modelname-#1of#2-#3.png}
    \caption{\texttt{#4}, \ #5}
    \label{fig:accuracyvskb_jpeg/all/\modelname_#3}
  \end{subfigure}%
}

\providecommand{\modelname}{}
\renewcommand{\modelname}{resnet34}

\providecommand{\shortcaption}{}
\renewcommand{\shortcaption}{%
  JPEG accuracy vs compressed output size (ResNet-34)%
}

\begin{figure}[H]
  \centering
  \textbf{\shortcaption}\par\medskip
  \begin{subfigure}[t]{\textwidth}
    \centering
    \includegraphics[width=\linewidth]
    {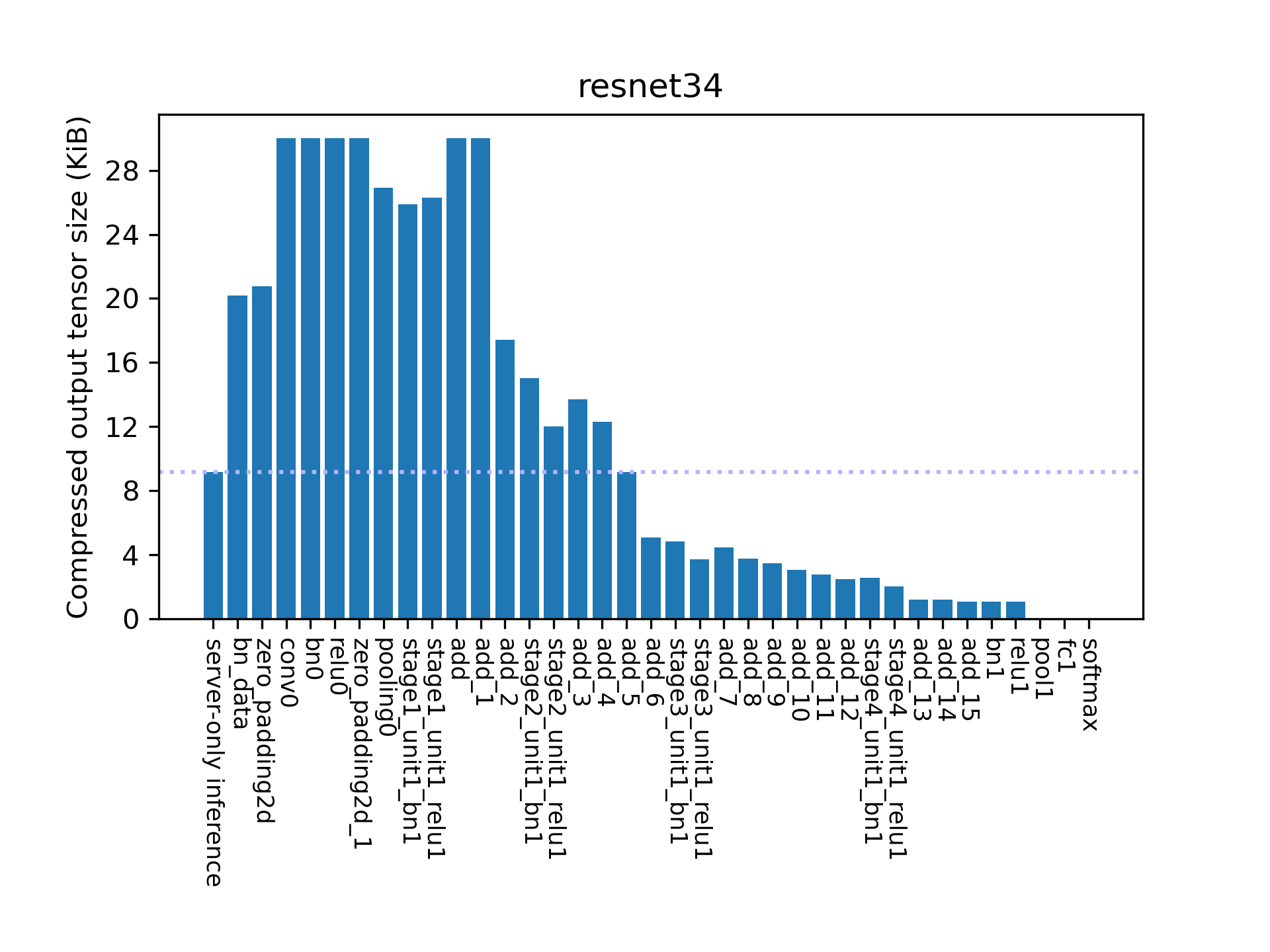}
    \caption{%
      compressed layer output size for $< 5\%$ drop in relative inference
      accuracy%
    }
    \label{fig:accuracyvskb_jpeg/all/\modelname_compressed}
  \end{subfigure}
\end{figure}

\begin{figure}[H]
  \ContinuedFloat
  \centering
  \begin{subfigure}[t]{\textwidth}
    \centering
    \includegraphics[width=\linewidth]
    {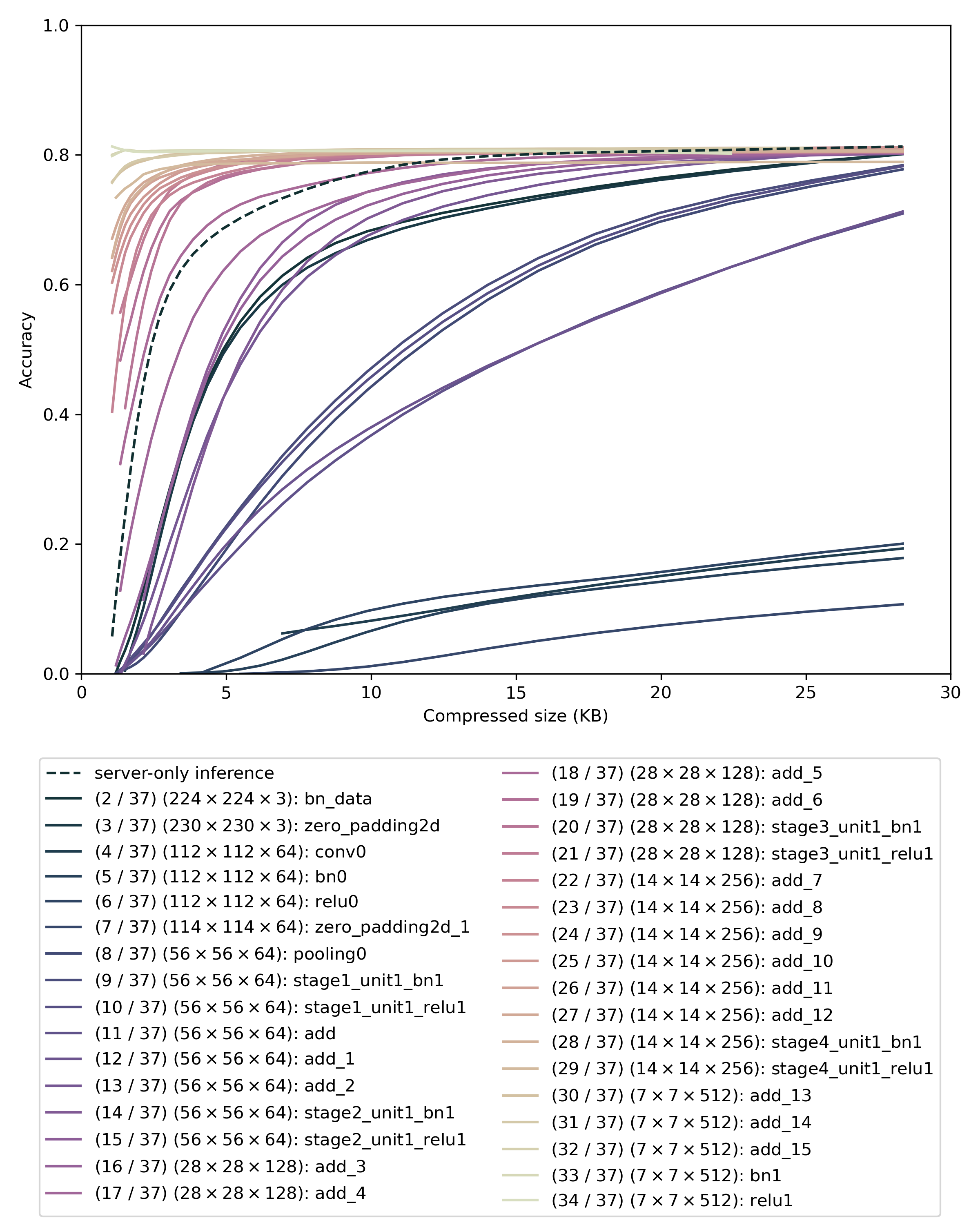}
    \caption{curves for all layers, smoothing applied}
    \label{fig:accuracyvskb_jpeg/all/\modelname_aggregate}
  \end{subfigure}
\end{figure}

\begin{figure}[H]
  \centering
  \ContinuedFloat
  \caption[\shortcaption]{%
    Top-1 image classification accuracy vs JPEG compressed frame output size
    across various layers of ResNet-34.%
  }
  \label{fig:accuracyvskb_jpeg/all}
\end{figure}

\providecommand{\fig}[5]{}
\renewcommand{\fig}[5]{%
  \begin{subfigure}[t]{.5\textwidth}
    \centering
    \includegraphics[width=\linewidth]
    {img/accuracyvskb/jpeg2000_uniquant256/\modelname/\modelname-#1of#2-#3.png}
    \caption{\texttt{#4}, \ #5}
    \label{fig:accuracyvskb_jpeg2000/all/\modelname_#3}
  \end{subfigure}%
}

\providecommand{\modelname}{}
\renewcommand{\modelname}{resnet34}

\providecommand{\shortcaption}{}
\renewcommand{\shortcaption}{%
  JPEG 2000 accuracy vs compressed output size (ResNet-34)%
}

\begin{figure}[H]
  \centering
  \textbf{\shortcaption}\par\medskip
  \begin{subfigure}[t]{\textwidth}
    \centering
    \includegraphics[width=\linewidth]
    {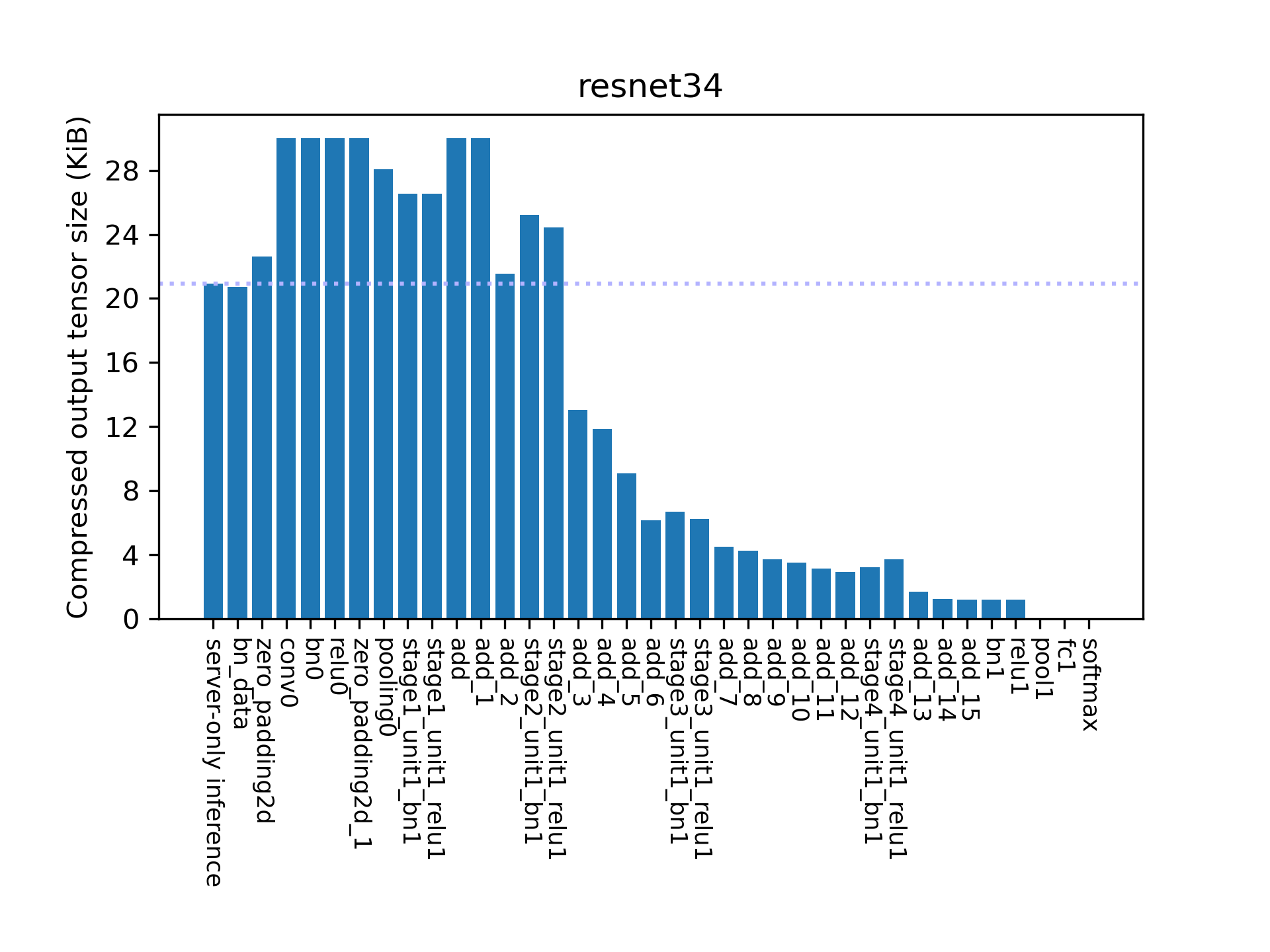}
    \caption{%
      compressed layer output size for $< 5\%$ drop in relative inference
      accuracy%
    }
    \label{fig:accuracyvskb_jpeg2000/all/\modelname_compressed}
  \end{subfigure}
\end{figure}

\begin{figure}[H]
  \ContinuedFloat
  \centering
  \begin{subfigure}[t]{\textwidth}
    \centering
    \includegraphics[width=\linewidth]
    {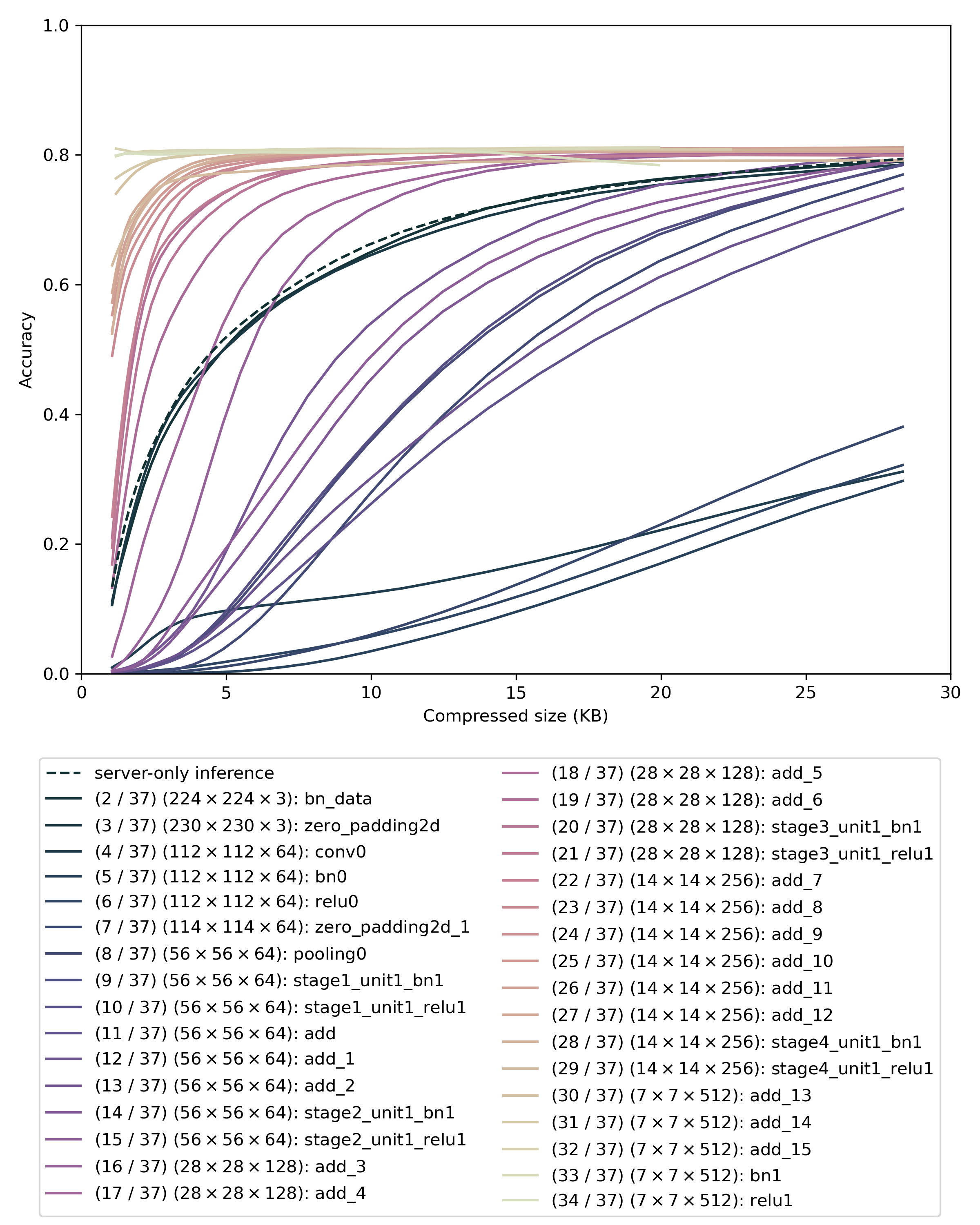}
    \caption{curves for all layers, smoothing applied}
    \label{fig:accuracyvskb_jpeg2000/all/\modelname_aggregate}
  \end{subfigure}
\end{figure}

\begin{figure}[H]
  \centering
  \ContinuedFloat
  \caption[\shortcaption]{%
    Top-1 image classification accuracy vs JPEG 2000 compressed frame output
    size across various layers of ResNet-34.%
  }
  \label{fig:accuracyvskb_jpeg2000/all}
\end{figure}

\end{document}